\documentclass[10pt,twocolumn,letterpaper]{article}

\usepackage[pagenumbers]{wacv} %

\usepackage{multirow}
\usepackage{float}
\usepackage{tikz}
\usetikzlibrary{patterns}
\usepackage{listings}
\usepackage{adjustbox}
\usepackage{seqsplit}

\PassOptionsToPackage{pagebackref,breaklinks,colorlinks}{hyperref}

\usepackage{orcidlink}

\usepackage{dblfloatfix}

\usepackage{makecell}
\usepackage{pifont}
\newcommand{\cmark}{\ding{51}}%
\newcommand{\xmark}{\ding{55}}%

\usepackage{etoolbox}
\providecommand{\nolinenumbers}{}  %
\AtBeginEnvironment{table}{\nolinenumbers}
\AtBeginEnvironment{table*}{\nolinenumbers}
\newif\ifsmalltabular \smalltabulartrue
\AtBeginEnvironment{tabular}{\ifsmalltabular\small\fi}
\pretocmd{\maketitle}{\smalltabularfalse}{}{}
\apptocmd{\maketitle}{\smalltabulartrue}{}{}

\newcommand{\precisioncolorbar}{%
  \begin{tikzpicture}[baseline=0pt]
    \node[left, font=\footnotesize] at (0,0.125) {GriTS\textsubscript{Con} cell error magnitude:\ \ \ };
    \shade[left color=white, right color=red, opacity=0.7]
      (0,0) rectangle (8,0.25);
    \draw[black, thin] (0,0) rectangle (8,0.25);
    \node[below, font=\footnotesize] at (0,-0.02) {0.0};
    \node[below, font=\footnotesize] at (4,-0.02) {0.5};
    \node[below, font=\footnotesize] at (8,-0.02) {1.0};
  \end{tikzpicture}%
}

\definecolor{annot@table}{rgb}{0.2,0.4,0.75}
\definecolor{annot@caption}{rgb}{0.1,0.7,0.5}
\definecolor{annot@footer}{rgb}{0.8,0.45,0.15}
\definecolor{annot@regular}{rgb}{0.45,0.55,0.7}
\definecolor{annot@spanning}{rgb}{0.7,0.15,0.55}
\definecolor{annot@prh}{rgb}{0.8,0.65,0.1}

\newcommand{\annotlegenditem}[2]{%
  \begin{tikzpicture}[baseline={([yshift=-.5ex]current bounding box.center)}]
    \fill[white] (0,0) rectangle (0.3,0.3);
    \draw[#1, line width=0.8pt] (0,0) rectangle (0.3,0.3);
    \clip (0,0) rectangle (0.3,0.3);
    \foreach \i in {-4,-3,...,4} {
      \draw[#1, line width=0.3pt, opacity=0.5] (\i*0.08,0) -- ++(0.3,0.3);
    }
  \end{tikzpicture}\,%
  {\footnotesize #2}%
}

\newcommand{\annotlegend}{%
  \par\vspace{4pt}%
  \centering
  \annotlegenditem{annot@table}{Table}\hspace{8pt}%
  \annotlegenditem{annot@caption}{Caption}\hspace{8pt}%
  \annotlegenditem{annot@footer}{Footer}\hspace{8pt}%
  \annotlegenditem{annot@regular}{Non-spanning cell}\hspace{8pt}%
  \annotlegenditem{annot@spanning}{Spanning cell}\hspace{8pt}%
  \annotlegenditem{annot@prh}{Projected row header}%
  \par
}

\definecolor{wacvblue}{rgb}{0.21,0.49,0.74}
\hypersetup{allcolors=wacvblue}

\def\wacvPaperID{180} %
\def\confName{WACV}
\def\confYear{2027}

\title{PubTables-v2: A new large-scale dataset for full-page and multi-page table extraction}

\author{Brandon Smock \qquad Maxim Sokolov \qquad Libin Liang \qquad Valerie Faucon-Morin \\
Amrit Ramesh \qquad Tayyibah Khanam \qquad Maury Courtland\\
Kensho Technologies\\
{\tt\small \{brandon.smock,maury.courtland\}@kensho.com}
}

\begin{document}
\maketitle
\begin{abstract}
Table extraction (TE) is a key challenge in document understanding.
Traditional approaches detect tables first, then recognize their structure.
Recently, interest has surged in developing methods, such as vision-language models (VLMs), to extract tables directly in their full page or document context.
However, a lack of annotated data has made progress difficult to demonstrate.
To address this, we create a new large-scale dataset, PubTables-v2.
PubTables-v2 unifies TE across various levels of surrounding context and, notably, is the first benchmark for multi-page TE.
Our evaluations reveal that while current frontier models strongly outperform ($+0.354\ \textrm{GriTS}_\textrm{Con}$) small models on the most complex task (full-document multi-page TE), this gap can be closed or even \emph{reversed} ($-0.056\ \textrm{GriTS}_\textrm{Con}$) on narrower tasks (cropped table extraction) with targeted training.
Data is available at \url{https://huggingface.co/datasets/kensho/PubTables-v2}.
\end{abstract}

\section{Introduction}
\label{sec:intro}

Table extraction (TE) has a long history as one of the most-studied problems in visual document understanding (VDU).
Traditionally, TE approaches have taken a document page as input, detected each individual table, and then processed each detected table separately \cite{schreiber2017deepdesrt, prasad2020cascadetabnet, smock2022pubtables, nassar2022tableformer}.

Although effective in many cases, this paradigm has a few potential drawbacks.
Commonly-cited is the complexity of developing and maintaining multi-stage extraction pipelines \cite{nassar2025smoldocling}.
Furthermore, important context could be lost as the table or portions of the table are processed in isolation.
This has sparked interest in developing methods that can extract tables directly within their full page context \cite{gemelli2023cte}.

\begin{figure}[tb]
    \centering
    \includegraphics[width=0.99\linewidth]{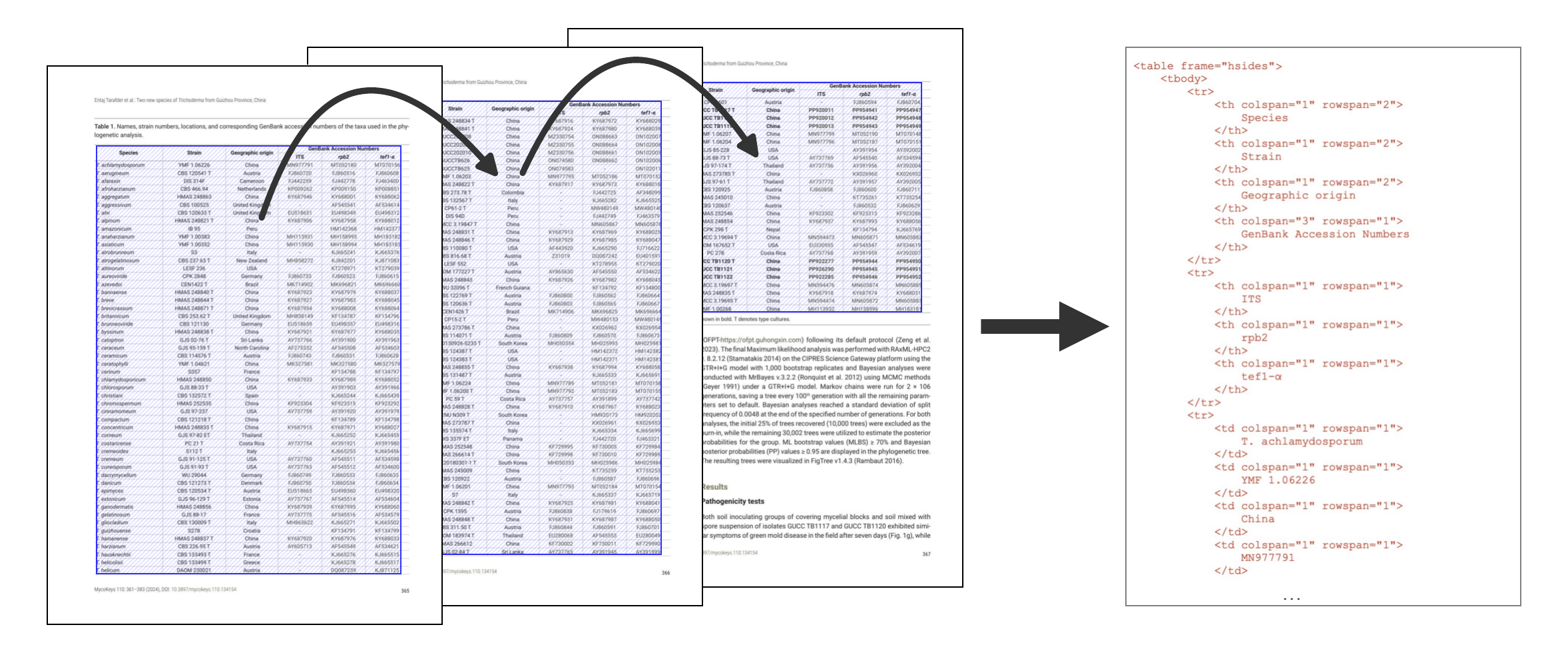}
    \caption{Among the more than 500k tables in PubTables-v2 are 9,492 multi-page tables, by far the largest set to date. PubTables-v2 includes tables spanning up to 13 pages.}
    \label{fig:multi_page_tables}
\end{figure}

Recently, auto-regressive models, ranging from frontier multimodal large language models (MLLMs)\footnote{Frontier MLLMs include Claude Opus 4.6 (Anthropic), GPT-5.4 (OpenAI), and Gemini 3.1 Pro (Google), accessible via commercial APIs.} to smaller specialized vision-language models (VLMs) \cite{bai2025qwen2, steiner2024paligemma, team2025granite, nassar2025smoldocling}, have emerged as a potential solution for end-to-end document parsing and \emph{contextualized} table extraction.
However, it remains standard practice to evaluate table structure recognition (TSR) performance solely on isolated cropped tables, rather than within a full page or document.

A key issue is the lack of established benchmarks and metrics for end-to-end TE.
Gemelli \etal~\cite{gemelli2023cte} proposed combining PubLayNet  \cite{zhong2019publaynet} document layout annotations with PubTables-1M \cite{smock2022pubtables} TSR annotations to create a dataset for page-contextualized table extraction.
However, the resulting dataset contains only 35k tables and does not maintain the train/test split of PubTables-1M, which creates the issue of data leakage.
Standard TSR metrics such as TEDS \cite{zhong2020image} and GriTS \cite{smock2023grits} assume a known one-to-one correspondence between predicted and ground truth tables, which breaks down when evaluating at the page and document level.

\begin{figure*}[tb]
    \centering
    \setlength{\fboxsep}{0pt}\setlength{\fboxrule}{0.4pt}
    \begin{tabular}{@{}c@{\hspace{2pt}}c@{\hspace{2pt}}c@{\hspace{2pt}}c@{}}
    \fbox{\includegraphics[height=0.325\textwidth]{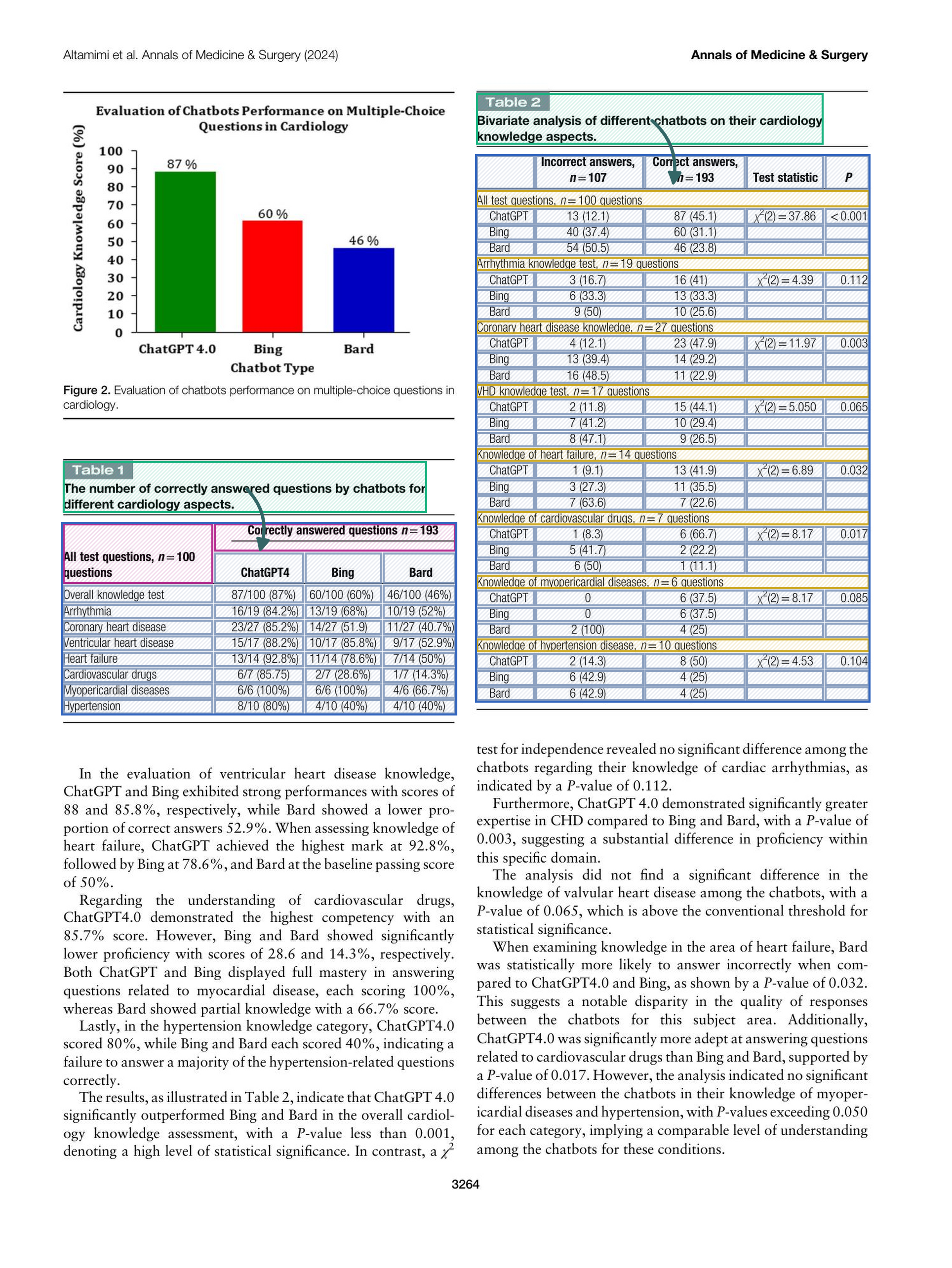}} &
    \fbox{\includegraphics[height=0.325\textwidth]{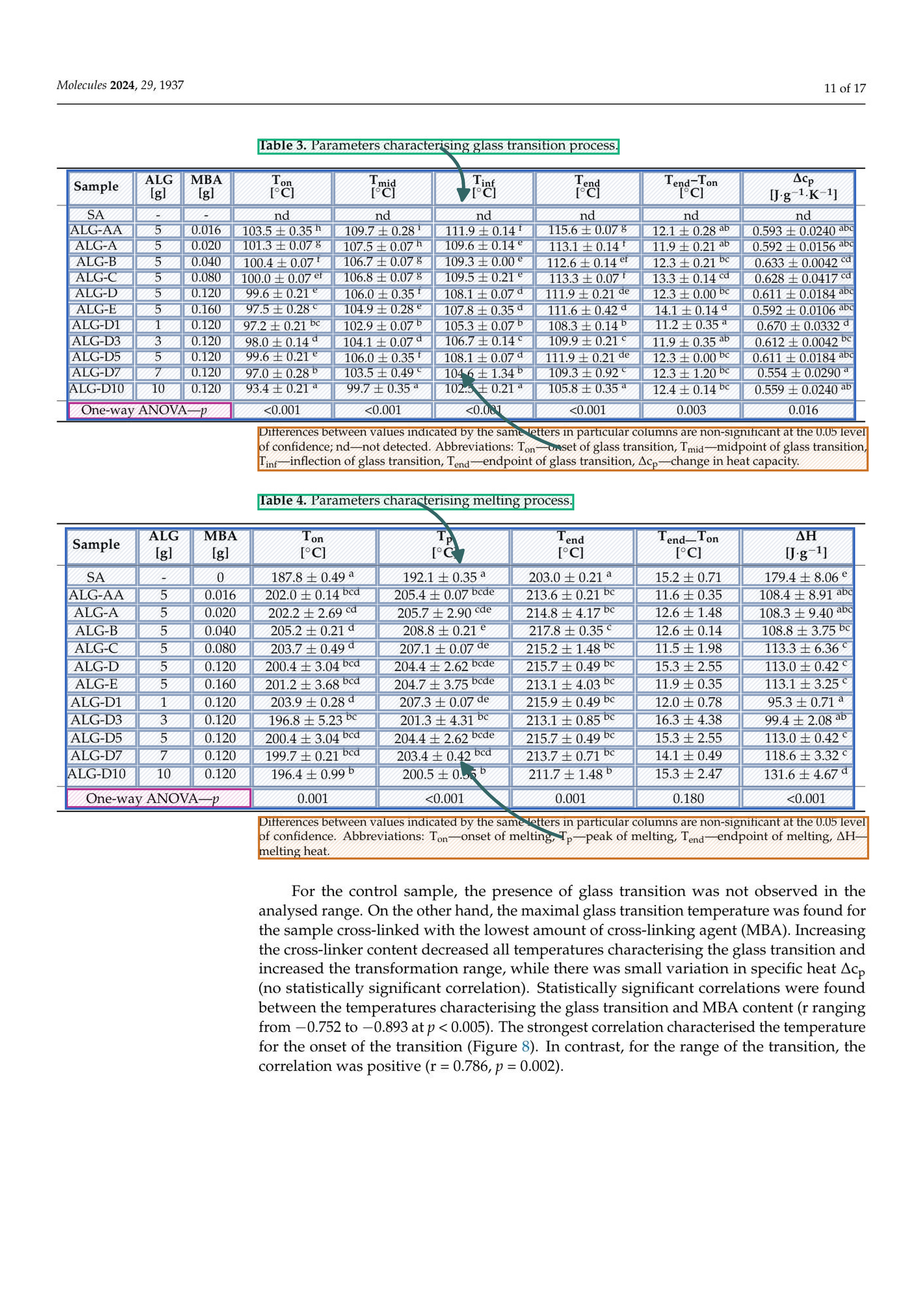}} &
    \fbox{\includegraphics[height=0.325\textwidth]{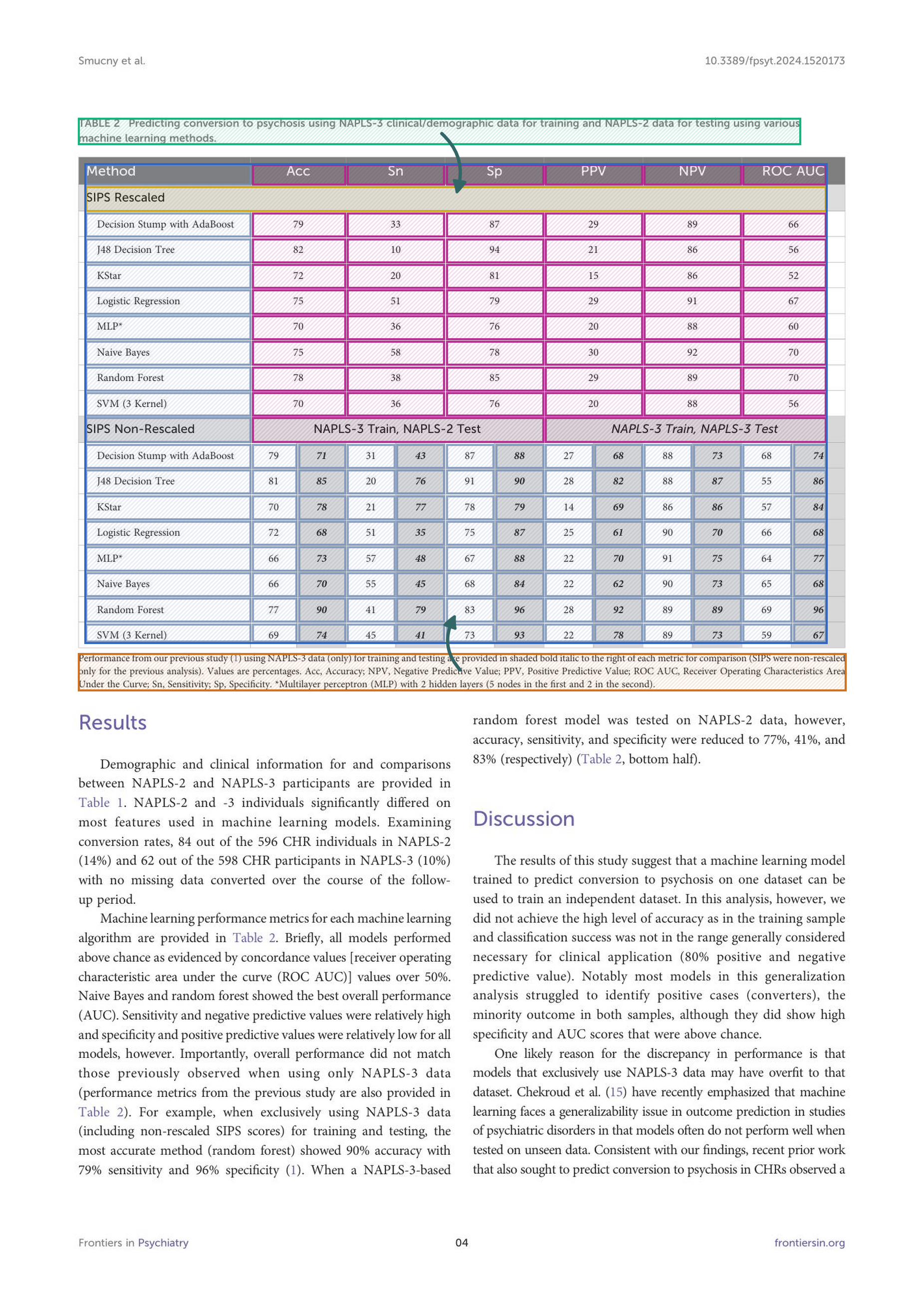}} &
    \fbox{\includegraphics[height=0.325\textwidth]{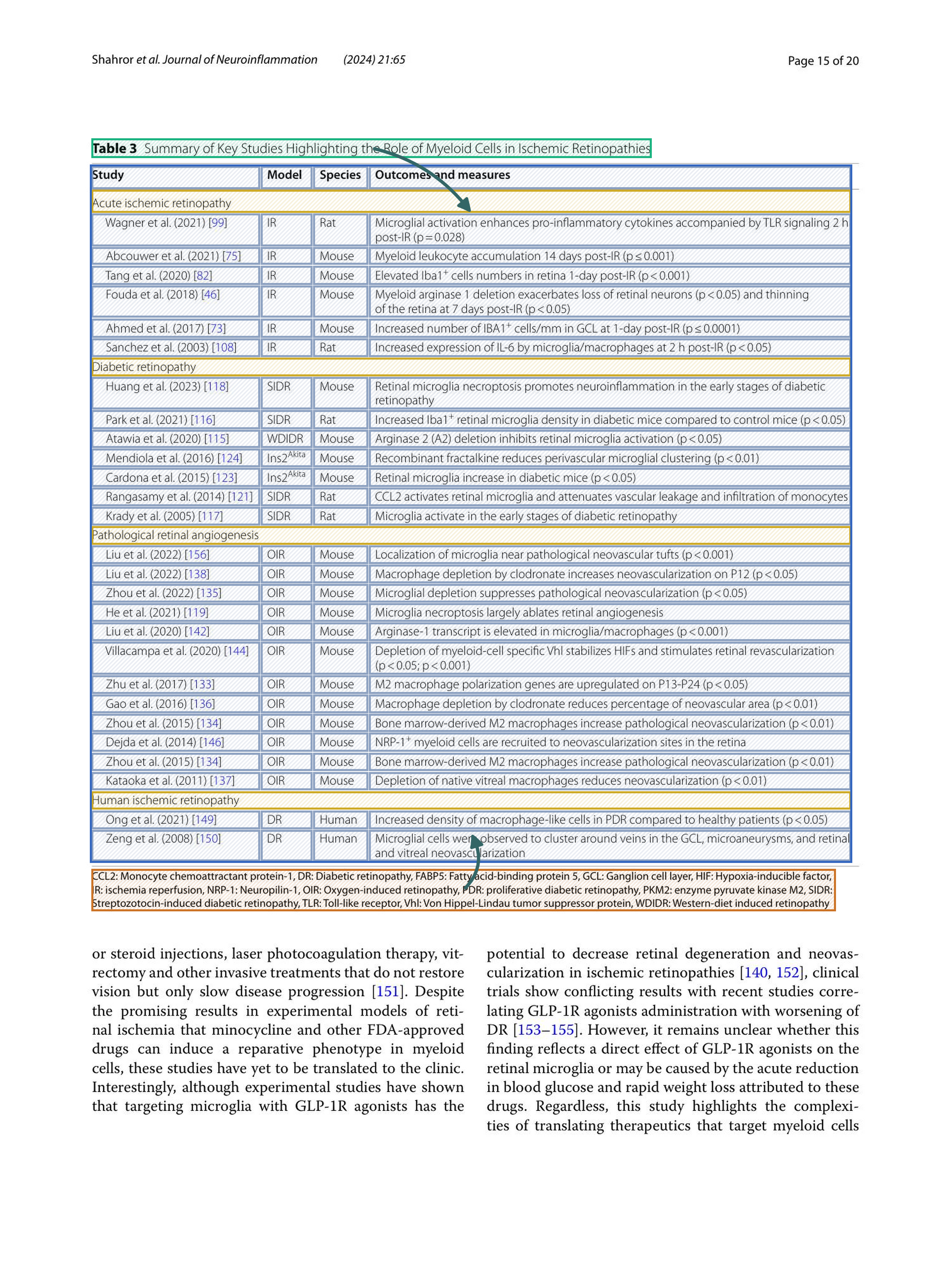}} \\
    \end{tabular}
    \annotlegend
    \caption{The PubTables-v2 Single Pages subset contains 548,414 tables in a full-page context. Above are four pages annotated with bounding boxes for table structure plus hierarchical relationships annotated with arrows between tables, their captions, and their footers.}
    \label{fig:annotation_quality_combined}
\end{figure*}

Another aspect of document parsing is inferring hierarchical or other relationships between page elements \cite{rausch2021docparser,ma2023hrdoc,wang2023graphical}.
However, no large-scale dataset exists that connects tables with both their captions and footers.

Beyond single pages, parsing content across multiple pages is an open area of study.
In visual question-answering (VQA) there is significant interest in developing methods that can answer questions requiring evidence from multiple pages \cite{ma2024mmlongbench}.
However, multi-page document parsing and multi-page table extraction are nascent areas of research.
In fact, no large dataset exists for multi-page table extraction, preventing this topic with significant real-world interest from being explored within the research literature.

Even well-established datasets for TSR \cite{zhong2020image, zheng2021global, smock2022pubtables} might benefit from new examples.
These datasets were established when models and the amount of input and output they could process were significantly smaller than they are today.
Current model pre-training practices have also introduced a much greater possibility for data leakage over time in long-standing datasets \cite{ramos2025data}.

Our goal in this work is to push the boundaries of table extraction with a new dataset and evaluation methodology addressing all the previously mentioned challenges.
Among our contributions:
\begin{itemize}
    \item We create PubTables-v2, a large-scale dataset supporting challenging aspects of table extraction at three levels of context: cropped tables, single pages, and full documents. It is the first large-scale benchmark for page-level and document-level table extraction, containing 548,414 tables in their full-page context with bounding boxes and hierarchical connections to captions and footers.
    \item We introduce the first large dataset of multi-page tables, with 9,492 instances spanning up to 13 pages, including over 200 tables that span at least 5 pages and 17 tables that span 10 pages or more.
    \item We propose extensions of standard TSR metrics---GriTS, TEDS, and exact match accuracy---to handle the most general case for page and document-level TE, where there can be multiple predicted and ground truth tables with no correspondence between them.
    \item We evaluate a broad range of models---from small trained narrow-task models to frontier MLLMs---for the first time across different levels of document context. Our results reveal the relative strengths of current model types, with frontier MLLMs' context advantage proving decisive for complex multi-page extraction but smaller models capable of achieving competitive or even superior performance on narrower tasks.
    \item We develop the first data-driven approach to cross-page table continuation prediction and show that merging of tables across pages significantly improves document-level TE performance for single-page-context models.
\end{itemize}

\section{Related Work}
\label{sec:related_work}

\begin{table}[tb]
\caption{\textbf{Comparison of long and wide cropped tables in document TSR datasets.}}
\centering
\begin{tabular}{l r r r}
\toprule
\textbf{Dataset Name} & \textbf{Total} & \textbf{Long} & \textbf{Wide} \\
\midrule
FinTabNet & 113k & 7,604 & 477 \\
PubTables-1M & 948k & 72,570 & 31,987 \\
\midrule
\textbf{PubTables-v2} & 136k & 100,781 & 37,106 \\
\bottomrule
\end{tabular} 
\label{tab:tsr_datasets}
\end{table}

\begin{table*}[tb]
\caption{\textbf{Comparison of datasets for page-level and document-level table extraction.} Page-Level Tables: tables annotated with structure in full-page context (not cropped). Pages w/ Structure: pages with $\geq$1 such table. No Missing Tables: whether all tables on annotated pages are guaranteed to be included (\cmark = guaranteed; \xmark = known missing; $\triangle$ = unlikely but not guaranteed). PubTables-v2 is the first dataset to provide page-level table structure annotations at scale with a completeness guarantee, and the first with multi-page tables.}
\centering
\begin{tabular}{l r r r r r c c}
\toprule
\textbf{Dataset} & \textbf{Year} & \textbf{\makecell{Pages w/\\Structure}} & \textbf{\makecell{Page-Level\\Tables}} & \textbf{\makecell{Full Docs\\w/ Structure}} & \textbf{\makecell{Multi-Page\\Tables}} & \textbf{\makecell{No Missing\\Tables}} & \textbf{\makecell{Cell\\Bbox}} \\
\midrule
ICDAR-2013 \cite{gobel2013icdar} & 2013 & 206 & 250 & 0\textsuperscript{\S} & 0 & \cmark & \cmark \\
ICDAR-2019 cTDaR \cite{gao2019icdar} & 2019 & 248 & 335 & 0 & 0 & \cmark & \cmark \\
FinTabNet \cite{zheng2021global} & 2021 & 76,077 & 112,887 & 0 & 0 & \xmark & \cmark \\
CTE \cite{gemelli2023cte} & 2023 & $<$35,000 & ${\sim}$35,000 & 0 & 0 & $\triangle$\textsuperscript{\ddag} & \cmark \\
OmniDocBench v1.5 \cite{ouyang2025omnidocbench} & 2025 & 458 & 665 & 0 & 0 & \xmark & \xmark \\
\midrule
PubTables-1M \cite{smock2022pubtables} & 2021 & 0\textsuperscript{\dag} & 0\textsuperscript{\dag} & 0 & 0 & ---\textsuperscript{\dag} & ---\textsuperscript{\dag} \\
\textbf{PubTables-v2} & \textbf{2026} & \textbf{467,541} & \textbf{548,414} & \textbf{9,172} & \textbf{9,492} & \cmark & \cmark\textsuperscript{\P} \\
\bottomrule
\multicolumn{8}{l}{\footnotesize{\textsuperscript{\dag}PubTables-1M provides page images with table detection annotations; structure annotations are for cropped table images only.}} \\
\multicolumn{8}{l}{\footnotesize{\textsuperscript{\ddag}Annotation procedure makes missing table annotations likely to be rare but does not guarantee no missing.}} \\
\multicolumn{8}{l}{\footnotesize{\textsuperscript{\S}Included PDFs appear to all be document excerpts, not full documents.}} \\
\multicolumn{8}{l}{\footnotesize{\textsuperscript{\P}Cell bounding boxes are not included for multi-page tables.}} \\
\end{tabular}
\label{tab:page_level_datasets}
\end{table*}

\subsection{Table Extraction Datasets}

Datasets for document-based table extraction typically separately annotate whole pages for table detection (TD) and cropped images of detected tables for table structure recognition (TSR).
Some prior datasets attempt to label table structures within their full-page context.
However, it has been difficult to scale these datasets while maintaining quality. \cref{tab:page_level_datasets} provides a comprehensive comparison, highlighting gaps in scale, annotation completeness, multi-page tables, and document-level coverage among existing page-level datasets.

FinTabNet \cite{zheng2021global} annotates 113k tables with structure information within 76k pages.
However, many tables within these pages are missing from the annotations.
Gemelli \etal~\cite{gemelli2023cte} propose a page-level table extraction dataset.
As noted previously, however, this dataset contains only 35k tables and models trained on it cannot also be trained on PubTables-1M \cite{smock2022pubtables} without data leakage, because the authors do not maintain the train/test split of PubTables-1M.

PubTables-1M contains 575k pages and uses strict quality control to prevent missing table annotations at the page level.
However, page images are only annotated for table detection, not structure recognition, which is annotated for cropped table images only.
Similarly, PubTabNet \cite{zhong2020image} provides structure annotations only for cropped tables.
More recently, OmniDocBench \cite{ouyang2025omnidocbench} includes tables annotated with structure within full pages as part of a broader document parsing benchmark.
However, it contains only 665 tables across 458 pages, does not guarantee annotation completeness, and does not contain cell-level bounding boxes.
To our knowledge, no large-scale dataset with strict quality control annotates table structure at the page or full-document level.
Furthermore, we are not aware of any published dataset that contains table recognition annotations for multi-page tables.

\subsection{Hierarchical Document Parsing}

Tables situated within documents often require their surrounding context to be fully understood.
Related to this is the problem of document layout analysis (DLA) \cite{zhong2019publaynet, pfitzmann2022doclaynet}, whose goal is to parse documents at a high-level into meaningful semantic units.

To fully understand page elements in context, however, requires recognizing the relationships between them.
DocParser \cite{rausch2021docparser} is one of the first attempts to recognize the full document relationship hierarchy.
However, creating high-quality data at this level of detail remains challenging.
The dataset used by DocParser for training and evaluation contained only 362 documents.
Larger datasets \cite{ma2023hrdoc, xing2024dochienet} have followed, but are still relatively small in terms of the number of documents, tables, captions, and their relationships.
We are not aware of any previous large-scale dataset that annotates tables with bounding boxes and connects them to their captions and footers.

\subsection{Vision-Language Models}

Both frontier MLLMs and smaller document-focused VLMs \cite{steiner2024paligemma, team2025granite, nassar2025smoldocling, li2025dots, granitedocling2025, wei2025deepseek, wei2026deepseek} prominently feature table extraction as a core capability.
However, presently very little evaluation of these models on real documents exists beyond isolated, cropped tables, which requires a separate table detection stage to produce and does not address broader challenges like multi-page table recognition.
Thus, some of the key claimed capabilities of VLMs for document parsing have not yet been rigorously validated.
This suggests a need for new benchmarks and evaluation procedures for page and document-level table extraction.

\section{PubTables-v2}
\label{sec:pubtables_v2}

\begin{figure}[tb]
    \centering
    \includegraphics[width=0.98\linewidth]{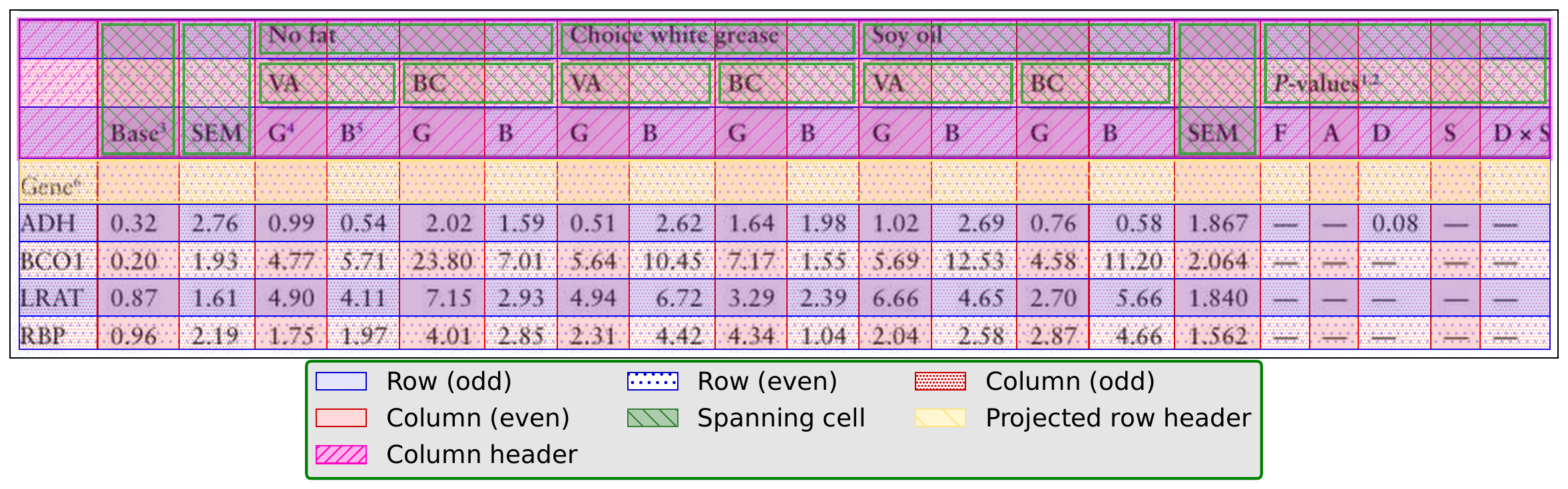}
    \caption{PubTables-v2 contains 136k long and wide cropped tables for TSR in the same format as PubTables-1M (see \cref{tab:tsr_datasets} for a comparison). Depicted above are structure annotations with bounding boxes for a wide table with 21 columns. }
    \label{fig:cropped_table}
\end{figure}

\begin{figure}[tb]
    \centering
    \includegraphics[width=\linewidth]{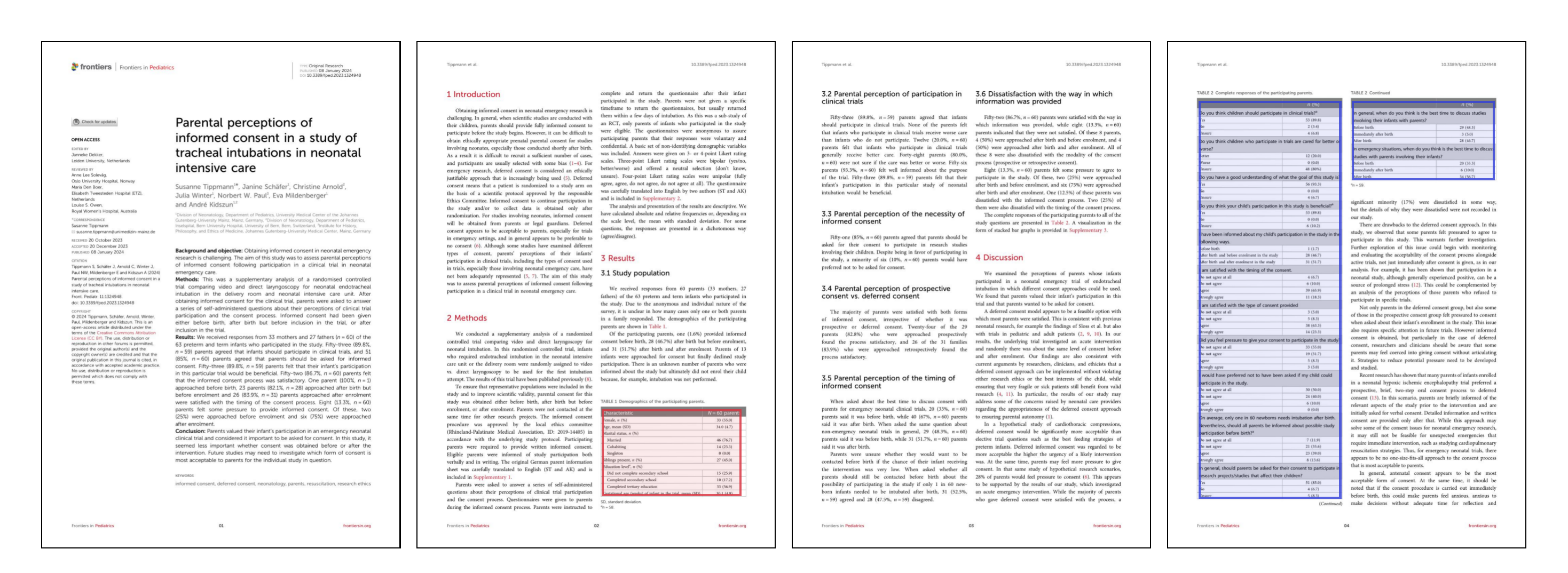}
    \caption{PubTables-v2 contains 9,172 full documents. Each document in this collection has at least one table that spans multiple pages or multiple columns within a page, as shown above. In this example, two tables in the document are highlighted: an isolated one in red and a single table split over two columns in blue.}
    \label{fig:full_document}
\end{figure}

\begin{table}[tb]
\caption{\textbf{Dataset statistics for PubTables-v2.} The data is organized into three collections by surrounding context: cropped tables, single pages, and full documents.}
\centering
{\footnotesize
\setlength{\tabcolsep}{3pt}
\begin{tabular}{@{}l r r r@{}}
\toprule
& \textbf{Cropped} & \textbf{Single} & \textbf{Full} \\
& \textbf{Tables} & \textbf{Pages} & \textbf{Documents} \\
\midrule
Samples & 136k & 468k & 9,172 \\
Tables & 136k & 548k & 25k \\
\quad Non-English/Multilingual & 348 & 682 & 34 \\
Pages & --- & 468k & 137k \\
\quad Two-column pages & --- & 37\% & 30\% \\
Documents & --- & --- & 9,172 \\
\midrule
Table rows (median) & 33 & 9 & 17 \\
Table rows (range) & 1--85 & 1--33 & 1--579 \\
Table columns (median) & 6 & 5 & 4 \\
Table columns (range) & 2--53 & 2--15 & 2--63 \\
Cells / table (median) & 152 & 42 & 66 \\
Spanning cells / table (median) & 8 & 4 & 6 \\
Tables with spanning cells & 80\% & 45\% & 48\% \\
\bottomrule
\end{tabular}}
\label{tab:dataset_statistics}
\end{table}

In this section, we describe PubTables-v2, our new large-scale, quality-controlled dataset for comprehensive page- and document-level TE.
PubTables-v2 is sourced from more than one million new articles published since PubTables-1M was released in 2021 and targets several challenging tasks not explored in the original dataset.

To create PubTables-v2, we adopt many of the same procedures described in Smock \etal~\cite{smock2022pubtables} that were used to create PubTables-1M.
Both datasets are created by leveraging XML (HTML) table annotations given by authors for every scientific article in PubMed.
Sequence alignment is used to match text from the HTML version with the text in the PDF version of the document, in order to locate each table and its cells within the PDF.

\subsection{Annotation Quality Control}
\label{sec:quality_control}

Because this alignment is not guaranteed to match the text correctly, it is important to identify failures and discard these in order to create a high-quality dataset.
After locating the cells through alignment, every cell's text within the PDF document's table is compared with the cell's text given in the original HTML annotation created by the PubMed article's author.
For a PDF table annotation to be considered high-quality, the mean normalized edit distance between the cells' text in the PDF and HTML must be less than a small threshold, $\alpha$.
Otherwise, it is discarded.

In PubTables-1M, $\alpha = 0.05$.
For PubTables-v2, we make this even more strict with $\alpha = 0.02$.
We also introduce a second threshold, $\beta$, such that the maximum normalized edit distance for all cells cannot exceed $\beta = 0.2$.
These strict conditions are a consistency verification step that ensures that for every table in the dataset, its PDF annotation agrees with the original author-supplied HTML annotation while forgiving minor differences like hyphenation at line breaks within a PDF.

For tables in PubTables-v2 with full page context (\ie the Single Pages and Full Documents collections), if any table originally annotated by the article's author does not pass this quality control step following alignment, \emph{the entire article} is discarded from the dataset.
This ensures that no page included in PubTables-v2 has any tables missing from the annotations.

To further improve the quality of annotations, we also adopt the table structure canonicalization scheme first proposed in PubTables-1M.
It has been shown that canonicalized annotations significantly improve labeling consistency and greatly improve the performance of TSR models trained on them \cite{smock2021tabletransformer, smock2023aligning}.
Finally, we adopt the improved column header annotation correction scheme proposed after the original PubTables-1M dataset by Smock \etal~\cite{smock2023aligning}.
This addresses the fact that PubMed authors do not always fully annotate the column headers of their tables.

\subsection{Collections}

PubTables-1M came in two collections: one with pages annotated for the TD task, and another with cropped table images annotated for TSR.
In PubTables-v2, we no longer decompose the table extraction problem into separate subtasks.
Rather, we consider end-to-end TE in different contexts.
By \emph{context} we mean how much of the surrounding source document is provided along with the table.
These different collections, each of which corresponds to a unique table context, are: 1) Cropped Tables, 2) Single Pages, and 3) Full Documents.
A detailed breakdown of each collection, including language, table structure, and page layout statistics, is given in \cref{tab:dataset_statistics}.

Addressing TE from each of these different contexts allows us to explore the unique challenges in each.
While the tables in each collection intentionally have unique characteristics that we detail in the following sections, each collection has the following in common: 1) each contains high-resolution images rendered from PDF documents; 2) for each image, all the words in the image, along with their bounding boxes, are extracted from the original PDF document and included as additional potential input; and 3) every table appearing in the input images is annotated with its bounding box location(s), its table structure, and the text content for every one of its cells, annotated in multiple formats and intended for use as ground truth output.
It is worth noting that each collection has the potential to support multiple fine-grained tasks within table extraction.
In the following sub-sections, we describe each collection in more detail.

\subsubsection{Cropped Tables}

Like PubTables-1M, PubTables-v2 contains a collection of cropped tables for TE, corresponding to the traditional TSR task.
In PubTables-1M, the most common table length is 7 rows, nearly 50\% of tables have 10 rows or fewer, and nearly 80\% of tables have 7 columns or fewer.
To explore more challenging scenarios, in PubTables-v2 we focus exclusively on long tables (which we define as having 30 rows or more) and wide tables (which we define as having 12 columns or more), for this task.

To create the collection of cropped tables, as previously mentioned, we gathered a new set of PubMed documents from 2023--2025.
Following quality control described in \cref{sec:quality_control} to remove low-quality annotations, our procedure yielded 135,578 tables with at least 30 rows or at least 12 columns.
An example of a wide table from this collection is shown in \cref{fig:cropped_table}.
We split these into three public sets---a train, a test, and a validation set---plus one additional hidden test set containing 5,804 samples (4.3\%) to detect data leakage in the future.

For training object detection-based models like TATR \cite{smock2021tabletransformer}, we include the same classes as the cropped tables collection of PubTables-1M: tables, columns, rows, column headers, projected row headers, and spanning cells.

\subsubsection{Single Pages}
\label{subsec:single_pages}

This collection contains tables annotated at the page level---in other words, it contains a collection of individual pages with annotations for every table on the page.
In addition to TSR annotations, we annotate bounding boxes for table captions and footers using the same sequence alignment procedure described for tables.

For training object detection models, there are 16 classes total: 8 object classes and their rotated counterparts for when objects of that type are rotated 90 degrees on the page (\ie as would occur within a rotated table).
The 8 object classes are: column, row, column header, projected row header, spanning cell, table, caption, and footer.
\cref{fig:annotation_quality_combined} contains an illustrated example of a page annotated for TE.

For training relation-prediction or image-to-graph models, we also include a set of relations between objects.
Given that there is no universally-established way to annotate relationships for page objects, we choose to do so hierarchically.
In this scheme, the table object is the parent and the other 7 object classes associated with it---such as its columns, rows, caption, and footers---are connected as children.
While some prior small datasets exist that connect tables to their captions, this is the first large-scale dataset we are aware of with hierarchical relationships between tables, their captions, and their footers.

In total, PubTables-v2 contains 468k individual pages annotated with 548k tables (see \cref{tab:dataset_statistics}).
Like in the cropped tables collection, we split pages into three public sets---a train, test, and validation set---plus one additional hidden test set for later use.

\begin{table}[t]
\caption{\textbf{Cropped table structure recognition (TSR).} Evaluation results on PubTables-v2 Cropped Tables. DT = document-extracted text (not OCR). The best zero-shot scores are \underline{\textit{italicized}}. Full results are in the Appendix (\cref{sec:appendix_full_results}).}
\centering
\begin{adjustbox}{max width=\columnwidth}
{\footnotesize
\begin{tabular}{l r r r r}
\toprule
\textbf{Model} & $\textbf{GriTS}_{\textbf{Top}}$ & $\textbf{GriTS}_{\textbf{Con}}$ & $\textbf{Acc}_{\textbf{Top}}$ & $\textbf{TEDS}$ \\
\midrule
SmolDocling-256M \cite{nassar2025smoldocling} & 0.2523 & 0.1923 & 0.1996 & 0.5928 \\
GraniteDocling-258M \cite{granitedocling2025} & 0.5482 & 0.5026 & 0.2065 & 0.6284 \\
Qwen2.5-VL-3b \cite{bai2025qwen2} & 0.6586 & 0.5305 & 0.1751 & 0.6672 \\
Granite-Vision-3.2-2b \cite{team2025granite} & 0.8714 & 0.7542 & 0.2155 & 0.6953 \\
DeepSeek-OCR 2 \cite{wei2026deepseek} & 0.7953 & 0.7429 & 0.2148 & 0.7148 \\
dots.ocr \cite{li2025dots} & 0.8343 & 0.8011 & 0.3190 & 0.7619 \\
\midrule
Claude Opus 4.6 & 0.9058 & 0.8952 & 0.3389 & 0.8512 \\
GPT-5.4 & 0.8842 & 0.8775 & 0.3124 & 0.8215 \\
Gemini 3.1 Pro & \underline{\textit{0.9354}} & \underline{\textit{0.9242}} & \underline{\textit{0.4518}} & \underline{\textit{0.8964}} \\
\midrule
TATR-v1.1-Fin \cite{smock2023aligning} + DT & 0.8985 & 0.8683 & 0.1405 & 0.8237 \\
TATR-v1.1-Pub \cite{smock2023aligning} + DT & 0.9608 & 0.9585 & 0.4867 & 0.9396 \\
\midrule
\multicolumn{5}{l}{\textbf{Trained on PubTables-v2}} \\
\quad TATR-v1.2-Pub + PaddleOCR \cite{cui2025paddleocr} & 0.9801 & 0.9007 & 0.6835 & 0.8809 \\
\quad TATR-v1.2-Pub + docTR \cite{doctr2021} & 0.9802 & 0.9286 & 0.6875 & 0.9231 \\
\quad TATR-v1.2-Pub + DT & \textbf{0.9806} & \textbf{0.9801} & \textbf{0.6892} & \textbf{0.9690} \\
\bottomrule
\end{tabular}}
\end{adjustbox}
\label{tab:experiment_tsr}
\end{table}

\subsubsection{Full Documents}
\label{sec:full_documents}

The third collection contains tables annotated at the document level.
Expanding from individual pages to full documents allows the annotation of tables that span multiple consecutive pages, which we refer to as multi-page tables.
PubTables-v2 is the first large-scale dataset for evaluating models on fully end-to-end document-level TE and the first dataset containing multi-page tables.

A key feature of our dataset is that every document in this collection contains at least one table that either continues across multiple pages, or in the case of two-column documents, across multiple page columns.
Each table is annotated with its structure, content, and the bounding boxes for each section of the table on each page it spans.
The first four pages of a document from this collection are shown in \cref{fig:full_document}.
While multi-page tables are the emphasis of this collection, for completeness we annotate all tables in the document, including single-page tables.
This enables several potential investigations into the effect of document-level context on general TE that would not be possible if we only annotated multi-page tables.
For more details on this collection, see the Appendix.

\section{Experiments}
\label{sec:experiments}

\subsection{Models}

For these experiments, we evaluate a broad range of model types, including frontier MLLMs (Claude Opus 4.6, GPT-5.4, Gemini 3.1 Pro) and smaller specialized VLMs \cite{bai2025qwen2,team2025granite, nassar2025smoldocling, li2025dots, granitedocling2025, wei2025deepseek, wei2026deepseek} trained for document parsing, all evaluated zero-shot on their core table extraction capabilities.
In addition, we report results for two small narrow-task models trained on PubTables-v2: TATR \cite{smock2021tabletransformer}, a 28M-parameter model for TSR, and its extension POTATR \cite{smock2026potatr}, a 29M-parameter model for page-level TE. These serve to demonstrate that PubTables-v2's annotations are high-quality and facilitate effective learning.
As these evaluations are only a snapshot at the time of data release, we also plan to establish a public leaderboard to measure ongoing progress, including whether fine-tuning of existing general-purpose models on PubTables-v2 improves performance.

\subsection{Metrics}
\label{sec:metrics}

Standard metrics for TSR performance include TEDS \cite{zhong2020image}, GriTS \cite{smock2023grits}, and table exact match accuracy (Acc), which is the percentage of tables for which the predictions and ground truth match exactly.
All of these metrics are designed to compare, for each input, exactly one predicted table matched with exactly one ground truth table.
However, in this paper we consider the most general case, where both the predictions and ground truth are \emph{sets} of tables---with no ordering and no correspondence given between them (\cref{fig:annotation_quality_combined})---and propose generalizations of GriTS, TEDS, and exact match accuracy to handle this case.

The generalizations of TEDS and GriTS use the Hungarian algorithm to determine the one-to-one match between the ground truth and predicted tables that maximizes each metric.
We aggregate the result as an F1 score, which penalizes poor precision in addition to recall for a given predicted set of tables.
Note that these metrics reduce to their original formulations in the case of one ground truth and one predicted table, as in traditional TSR, making them strict generalizations rather than replacements.
We use the generalized versions of GriTS and TEDS for all experiments.
For further details about the metrics, please see the Appendix.%

\subsection{Cropped Table Structure Recognition}

In this experiment, we evaluate models on their ability to recognize the structure of long and wide tables from the cropped tables collection of PubTables-v2.
We fine-tune \texttt{\seqsplit{TATR-v1.1-Pub}} on the train split of the cropped tables collection of PubTables-v2 to show the impact that further training has for improving model performance.
We refer to our fine-tuned model as \texttt{\seqsplit{TATR-v1.2-Pub}}.

Unlike the VLMs, TATR does not recognize text and requires its table structure predictions to be combined with text either extracted directly from the document or from a separate OCR model.
We compare TATR with text produced in three ways: from two OCR models (PaddleOCR \cite{cui2025paddleocr} and docTR \cite{doctr2021}), along with text extracted directly from the document (abbreviated DT).

\begin{figure*}[tb]
    \centering
    \setlength{\fboxsep}{0pt}\setlength{\fboxrule}{0.4pt}
    \begin{tabular}{@{}c@{\hspace{6pt}}c@{\hspace{6pt}}c@{\hspace{6pt}}c@{}}
    \fbox{\includegraphics[width=0.235\textwidth]{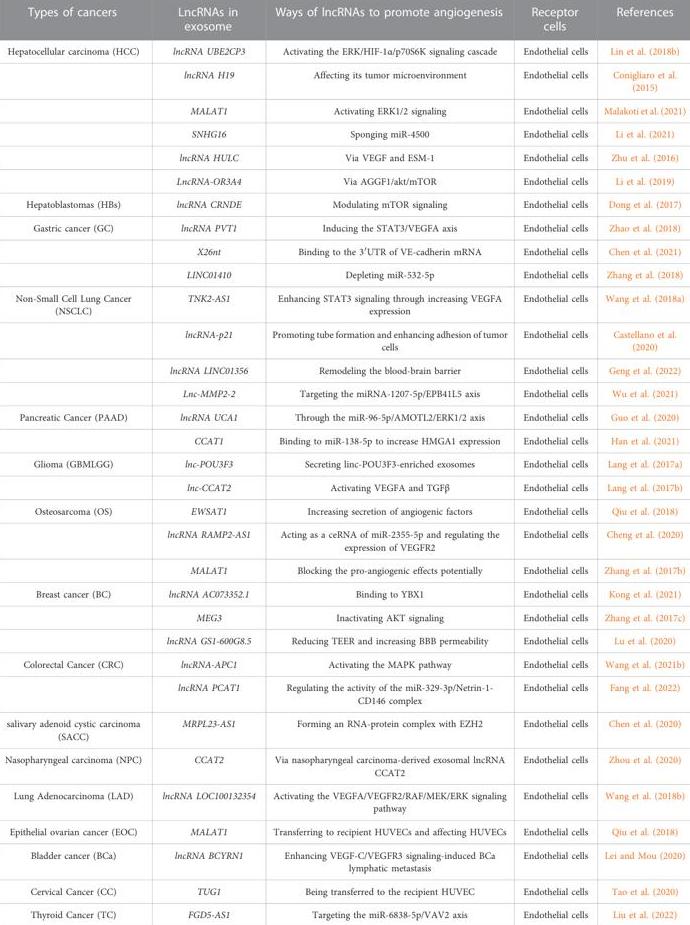}} &
    \fbox{\includegraphics[width=0.235\textwidth]{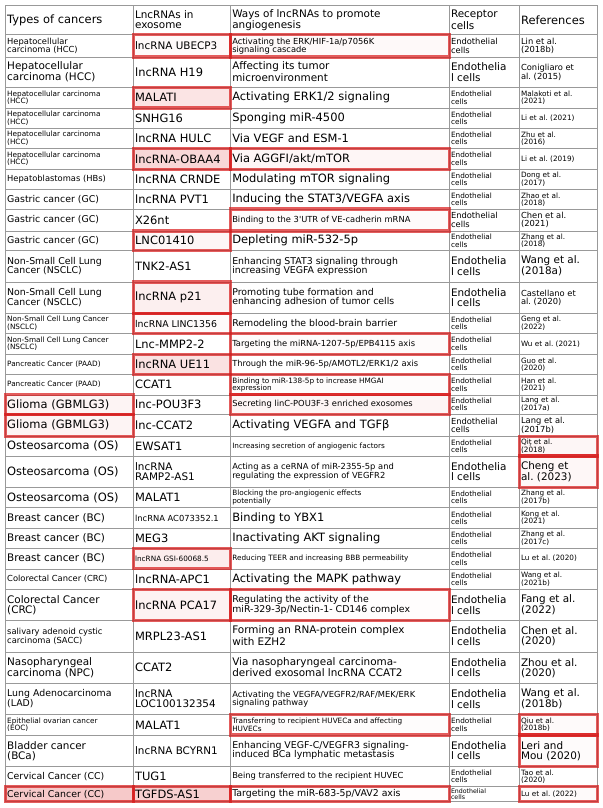}} &
    \fbox{\includegraphics[width=0.235\textwidth]{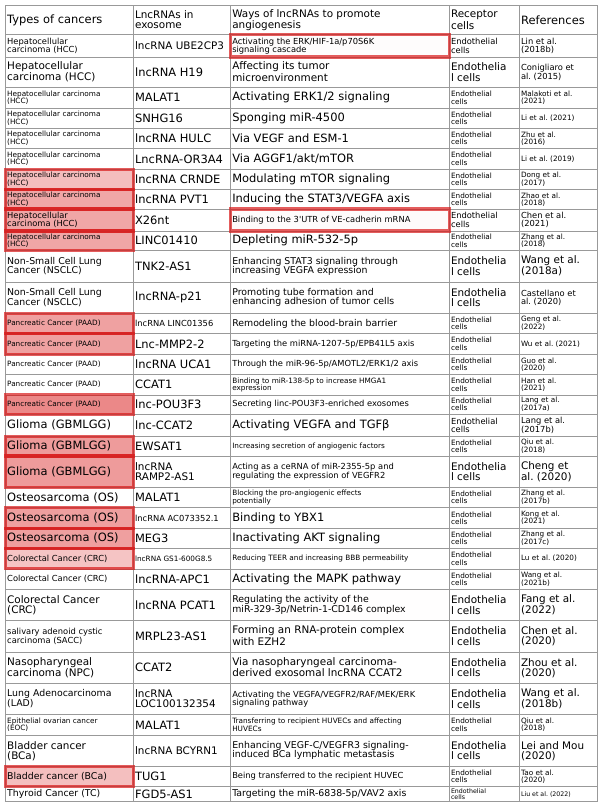}} &
    \fbox{\includegraphics[width=0.235\textwidth]{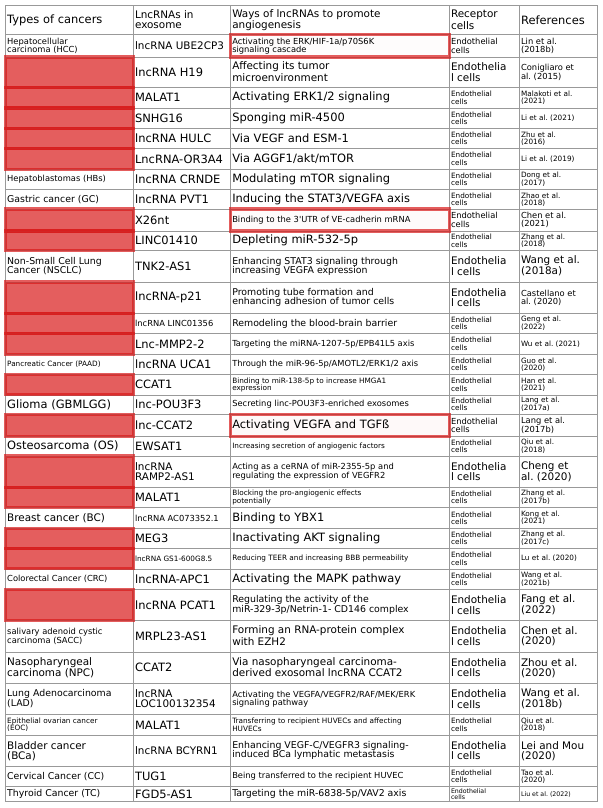}} \\
    {\footnotesize (a) Original table} &
    {\footnotesize (b) SmolDocling} &
    {\footnotesize (c) DeepSeek-OCR} &
    {\footnotesize (d) Gemini 3.1 Pro} \\
    \end{tabular}
    \par\vspace{4pt}
    \precisioncolorbar
    \caption{\textbf{Qualitative failure analysis} for table PMC10436220\_table\_0 (34 rows $\times$ 5 columns). Each cell in panels (b)--(d) shows the predicted text content, with the content of predicted spanning cells repeated at each spanned grid location; red-bordered cells contain errors, with background shading proportional to error magnitude (white = perfect match, red = complete mismatch). SmolDocling exhibits correct structure with minor character-level errors (e.g., $\alpha \to$ a). DeepSeek-OCR exhibits errors recognizing the true rowspans of the spanning cells in the first column. Gemini 3.1 Pro fails to recognize the spanning cells altogether, segmenting them into multiple content and blank cells.}
    \label{fig:precision_qualitative_1row}
\end{figure*}

The results of this experiment are given in \cref{tab:experiment_tsr}.
Of the smaller specialized VLMs, \texttt{\seqsplit{dots.ocr}} performs best, achieving a $\textrm{GriTS}_\textrm{Con}$ of 0.801.
Frontier MLLMs significantly outperform the smaller VLMs, with Gemini 3.1 Pro achieving a $\textrm{GriTS}_\textrm{Con}$ of 0.924.
Failure modes can vary widely across models, as we illustrate in
\cref{fig:precision_qualitative_1row}.

\texttt{\seqsplit{TATR-v1.2-Pub}} paired with the best OCR model, docTR, achieves a $\textrm{GriTS}_\textrm{Con}$ of 0.929---comparable to the best frontier MLLM.
When utilizing document text directly, \texttt{\seqsplit{TATR-v1.2-Pub}} achieves a $\textrm{GriTS}_\textrm{Con}$ of 0.980.
This cuts the error in half compared to \texttt{TATR-v1.1-Pub} (0.959 $\textrm{GriTS}_\textrm{Con}$), which shows the impact that fine-tuning on a large number of additional samples has on recognition performance for long and wide tables.

\subsection{Page-level Table Extraction}\label{sec:page_level_te}

\begin{table}[tb]
\caption{\textbf{Page-level table extraction.} Evaluation results on PubTables-v2 Single Pages. DT = document-extracted text (not OCR). The best zero-shot scores are \underline{\textit{italicized}}. Additional results are in the Appendix (\cref{sec:appendix_full_results}).}
\centering
\begin{adjustbox}{max width=\columnwidth}
\begin{tabular}{l r r r r}
\toprule
\textbf{Model} & $\textbf{GriTS}_{\textbf{Top}}$ & $\textbf{GriTS}_{\textbf{Con}}$ & $\textbf{Acc}_{\textbf{Top}}$ & $\textbf{TEDS}$ \\
\midrule
SmolDocling-256M & 0.2502 & 0.2177 & 0.2800 & 0.5250 \\
GraniteDocling-258M & 0.7252 & 0.6670 & 0.3794 & 0.6155 \\
Qwen2.5-VL-3b & 0.4672 & 0.3803 & 0.3651 & 0.2632 \\
Granite-Vision-3.2-2b & 0.8015 & 0.7480 & 0.4245 & 0.8092 \\
DeepSeek-OCR 2 & 0.8839 & 0.8386 & 0.5258 & 0.7263 \\
dots.ocr & 0.9241 & 0.8991 & \underline{\textit{0.6425}} & 0.8743 \\
\midrule
Claude Opus 4.6 & 0.9146 & 0.9106 & 0.4934 & 0.8945 \\
GPT-5.4 & 0.9208 & 0.9102 & 0.5461 & 0.8874 \\
Gemini 3.1 Pro & \underline{\textit{0.9500}} & \underline{\textit{0.9418}} & 0.6348 & \underline{\textit{0.9250}} \\
\midrule
\multicolumn{5}{l}{\textbf{Trained on PubTables-v2}} \\
\quad POTATR \cite{smock2026potatr} + PaddleOCR & 0.9614 & 0.8768 & 0.6953 & 0.8780 \\
\quad POTATR \cite{smock2026potatr} + docTR & 0.9606 & 0.8809 & 0.6724 & 0.8971 \\
\quad POTATR \cite{smock2026potatr} + DT & \textbf{0.9674} & \textbf{0.9647} & \textbf{0.7012} & \textbf{0.9516} \\
\bottomrule
\end{tabular}
\end{adjustbox}
\label{tab:page_level_TSR_results}
\end{table}

In this experiment, we evaluate models on their ability to extract tables at the page level.
In other words, models must take in an entire page and produce TSR output for all of the tables contained within the page.
Unlike the cropped tables experiment, which focuses exclusively on long and wide tables, the distribution of tables in this collection is more representative of PubMed as a whole.

Results are given in \cref{tab:page_level_TSR_results}.
Among the zero-shot models, frontier MLLMs significantly outperform smaller specialized VLMs, with Gemini 3.1 Pro achieving the highest $\textrm{GriTS}_\textrm{Con}$ of 0.942.
However, POTATR \cite{smock2026potatr} with text extracted from the document achieves $\textrm{GriTS}_\textrm{Con}$ of 0.965---outperforming the best frontier MLLM despite being a 29M-parameter model.
The gap between POTATR paired with traditional OCR (0.877--0.881) and POTATR with direct text (0.965) indicates that remaining errors are dominated by text quality rather than structure recognition.

\subsection{Document-level Table Extraction}
\label{sec:doc_level_te}

\begin{table}[tb]
\caption{\textbf{Document-level table extraction.} Evaluation results on PubTables-v2 Full Documents. Frontier MLLMs process full documents at once; smaller VLMs process each page individually; cross-page merging uses a ViT model to predict table continuations (\cref{sec:exp_cross_page_tables}). Full results with all metrics and models are in the Appendix (\cref{sec:appendix_full_results}).}
\centering
\begin{adjustbox}{max width=\columnwidth}
\begin{tabular}{l r r r r}
\toprule
\textbf{Model/Context} & $\textbf{GriTS}_{\textbf{Top}}$ & $\textbf{GriTS}_{\textbf{Con}}$ & $\textbf{Acc}_{\textbf{Top}}$ & $\textbf{TEDS}$ \\
\midrule
\multicolumn{5}{l}{\textbf{Full document}} \\
\quad Claude Opus 4.6 & 0.9116 & 0.9079 & \textbf{0.5799} & 0.8798 \\
\quad GPT-5.4 & 0.8865 & 0.8847 & 0.5038 & 0.8341 \\
\quad Gemini 3.1 Pro & \textbf{0.9365} & \textbf{0.9309} & 0.4854 & \textbf{0.8905} \\
\multicolumn{5}{l}{\textbf{Pages processed individually}} \\
\quad GraniteDocling-258M & 0.5260 & 0.4723 & 0.2732 & 0.4470 \\
\quad DeepSeek-OCR 2 & 0.4920 & 0.4527 & 0.3661 & 0.5176 \\
\quad dots.ocr & 0.6054 & 0.5768 & 0.3958 & 0.5876 \\
\multicolumn{5}{l}{\textbf{Pages with cross-page merging (\cref{sec:exp_cross_page_tables})}} \\
\quad dots.ocr + merging & 0.7772 & 0.7495 & 0.5059 & 0.7731 \\
\bottomrule
\end{tabular}
\end{adjustbox}
\label{tab:experiment_multi_page_table_extraction}
\end{table}

In this experiment, we evaluate models on their ability to extract tables at the document level.
This is a difficult task requiring models to perform well at several sub-tasks including extracting potentially multiple tables simultaneously in a full-page context, ignoring pages without tables, and recognizing when a single table is split into multiple parts that may span subsequent pages.

We evaluate models under two different settings.
In the first setting, a model is given all the pages of a document at once and must produce all the extracted tables from the document.
We evaluate frontier MLLMs in this setting.
Most smaller specialized VLMs, however, are limited to single-page inputs.
In the second setting, each page is input and processed by the model independently.
While the processing is done page-by-page, the evaluation remains at the document level, comparing all predicted tables with all ground truth tables for the entire document at once.
The results of this experiment are presented in \cref{tab:experiment_multi_page_table_extraction}.

Frontier MLLMs with full-document context perform strongly, with Gemini 3.1 Pro achieving $\textrm{GriTS}_\textrm{Con}$ of 0.931.
In contrast, smaller VLMs processing pages individually struggle significantly---the best, \texttt{dots.ocr}, achieves only 0.577.
This large gap highlights multi-page table recognition as a key challenge for page-level models and demonstrates the value of full-document context for this task.
For further analysis of model performance on this collection, please see the Appendix.

\subsection{Cross-page Table Continuation}\label{sec:exp_cross_page_tables}

\begin{figure}[tb]
    \centering
    \includegraphics[width=0.98\linewidth]{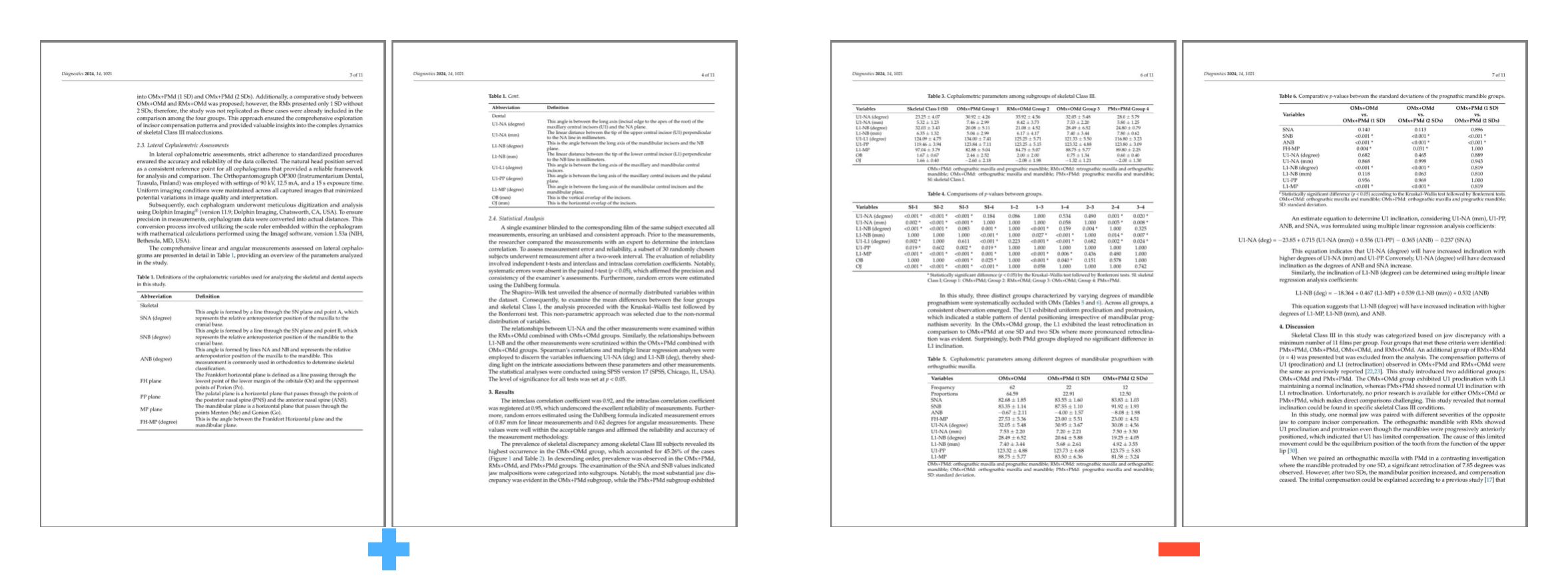}
    \caption{Two samples from the cross-page table continuation task. Left (\textbf{\textcolor{cyan}{+}}): a positive pair where the table continues onto the next page. Right (\textbf{\textcolor{red}{--}}): a negative pair. Both pairs are from the same document.}
    \label{fig:cross_page_table_continuation}
\end{figure}

\begin{table}[tb]
\caption{\textbf{Cross-page table continuation classification.} Results for models trained on PubTables-v2's continuation prediction task.}
\centering
\begin{tabular}{l r r r r}
\toprule
\textbf{Model} & \textbf{Recall} & \textbf{Precision} & \textbf{F1} & \textbf{AUC} \\
\midrule
ResNet-50 & 0.986 & 0.973 & 0.979 & 0.991 \\
ViT-B-16 & \textbf{0.995} & \textbf{0.987} & \textbf{0.991} & \textbf{0.996} \\
\bottomrule
\end{tabular}
\label{tab:experiment_cross_page_table_continuation}
\end{table}

To address the limitations of page-by-page processing, we propose a new task: cross-page table continuation prediction, where a model is given a pair of page images and must predict whether the last table on the first page continues onto the second.
PubTables-v2 contains 9,866 positive pairs of pages where a table on the first page continues onto the second page.
For negative samples, we focus only on contiguous pages, each with tables, that each come from the same document as a positive sample, limiting the potential for shortcut learning \cite{geirhos2020shortcut} to exploit pure stylistic differences.
This constraint yielded 5,964 negative pairs.
We illustrate a positive and a negative example in \cref{fig:cross_page_table_continuation}.

To train a model on this task, we frame it as binary classification, where the input is the two contiguous page images concatenated side-by-side.
We choose this framing as the simplest way to explore a novel data-driven approach to table continuation prediction.
We train two models, ResNet-50 \cite{he2016deep} and ViT-B-16 \cite{dosovitskiy2020image}.
The results are given in \cref{tab:experiment_cross_page_table_continuation}.
\texttt{ViT-B-16} outperforms \texttt{ResNet-50}, achieving a recall of 0.995 while maintaining a precision of 0.987.
The excellent performance of both models suggests that in PubTables-v2 there are strong visual cues to indicate whether a table on one page continues onto the next.

Using this result, we attempt to improve page-by-page document TE by using the ViT model to predict when to merge tables produced by \texttt{dots.ocr}.
When a continuation is predicted and the adjacent tables have matching column counts, we concatenate them vertically.
The result of this experiment is given in \cref{tab:experiment_multi_page_table_extraction}.
Model-based cross-page table merging leads to a significant boost in performance, improving $\textrm{GriTS}_{\textrm{Con}}$ from 0.577 to 0.750.
This demonstrates that PubTables-v2 can enable practical solutions to multi-page table extraction challenges that were previously unexplored due to a lack of data.

\section{Limitations}
\label{sec:limitations}

PubTables-v2 is sourced from tables in millions of PubMed scientific articles.
Our analysis shows that 99.85\% of these tables are in English, 0.13\% are non-English, and 0.02\% are mixed.
While diverse in many aspects (\cref{tab:dataset_statistics}), PubTables-v2 is not intended to replace other datasets that address additional variation in content, language, and appearance.

However, we note that domain limitation is what enables PubTables-v2 to scale to over 700k tables while maintaining quality.
A bootstrap analysis (\cref{sec:appendix_ranking_stability}) of test data subsets shows that at $N{=}500$, agreement with the full dataset ranking of adjacent models drops to ${\sim}76\%$.
This instability strongly suggests that scale is essential for reliable evaluation.
Thus, PubTables-v2's depth represents a useful trade-off that complements the breadth of multi-domain datasets.

\section{Conclusion and Future Work}
\label{sec:conclusion}

In this work, we introduced PubTables-v2, a new large-scale dataset that unifies TE across three levels of context.
Notably, PubTables-v2 is the first benchmark for multi-page table extraction, with tables spanning up to 13 pages.
Our evaluation across a broad range of model types and document contexts reveals a clear tension in the current state of the art: frontier MLLMs' ability to leverage full-document context gives them a decisive advantage on complex multi-page extraction, yet this advantage disappears as the task narrows, with small models trained on PubTables-v2 capable of achieving competitive or even superior performance.
To help address the context gap, we used PubTables-v2 to develop the first data-driven approach to cross-page table continuation prediction.
We found that image classifiers can predict table continuations with near-perfect accuracy and that model-based merging significantly improves document-level TE for single-page-context models.

Altogether, these findings suggest a number of important directions remain.
Future work could explore multi-page table recognition in more depth, including novel approaches, cost-efficient recognition, visual grounding at the cell level, and cross-domain generalization.
Another interesting direction is to create a dedicated multi-lingual evaluation by applying the same dataset construction pipeline to a larger sample of multi-lingual PubMed articles.
We hope our dataset enables others to build on our contributions and continue to push the boundaries in these directions.

{
    \small
    \bibliographystyle{ieeenat_fullname}
    \bibliography{main}
}

\clearpage

\section*{Appendix}

\appendix

\section{Licenses}

The PubTables-v2 dataset is open sourced under the Community Data License Agreement (CDLA) – Permissive, Version 2.0.
Models are available under the MIT license.

\section{Dataset}
\label{sec:appendix_pubtables_v2}

In this section we provide additional details, statistics, and examples for the PubTables-v2 dataset discussed in \cref{sec:pubtables_v2}.

\subsection{Format}
\label{sec:appendix_dataset_format}

To leverage existing model training and evaluation code \cite{smock2021tabletransformer}, we annotate the dataset for training object detection models in the same format as PubTables-1M \cite{smock2022pubtables}, which contains annotations in PASCAL VOC format.
We annotate the ground truth for table structure recognition (TSR) evaluation in three formats: 1) the grid (matrix) format used by the GriTS \cite{smock2023grits} metric, 2) HTML format used by the TEDS \cite{zhong2020image} metric, and 3) a list of cells in JSON format.
For training image-to-graph models, relations between objects are annotated in JSON format.
Annotations for multi-page tables and single-page split tables are also annotated in JSON format.
Please see the dataset README for more information.

\subsection{Multi-Part Tables}
\label{sec:appendix_pubtables_v2_multi_page}

\begin{figure*}[tb]
    \centering
    \includegraphics[height=66px]{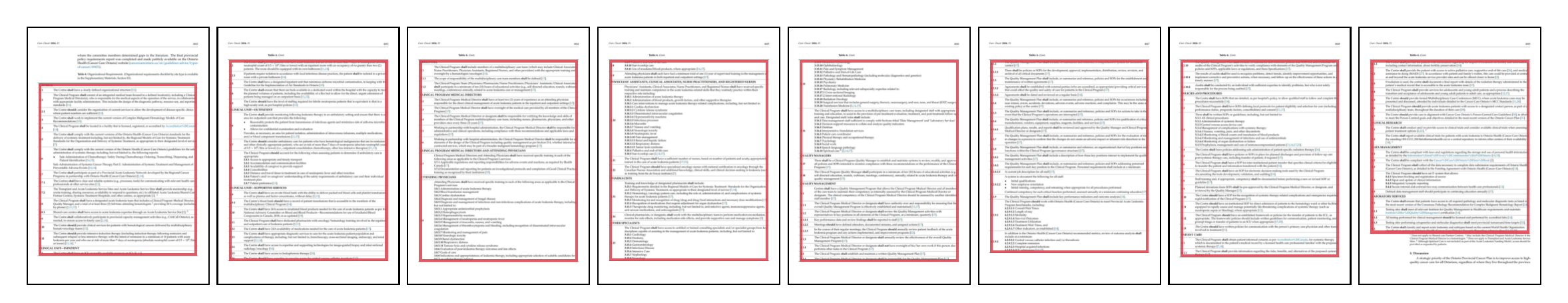} \\
    \includegraphics[height=66px]{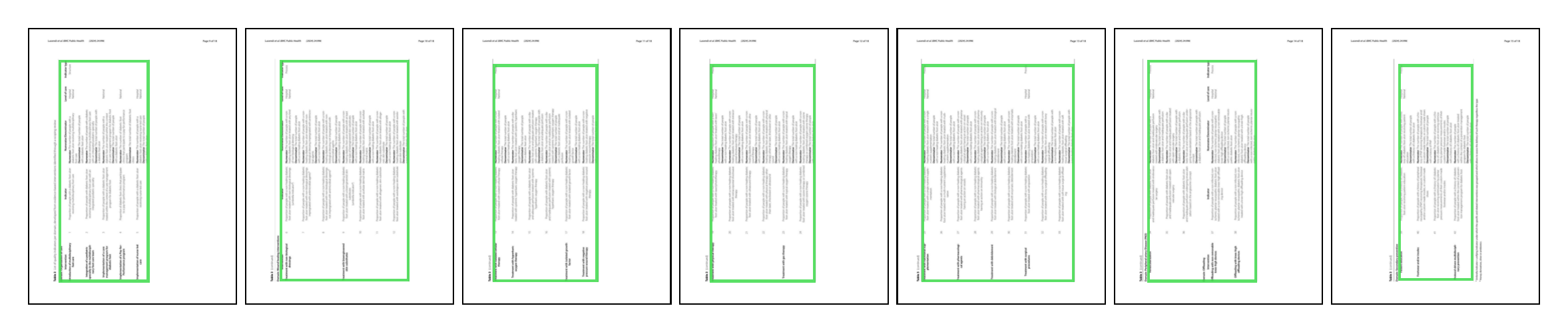} \\
    \includegraphics[height=66px]{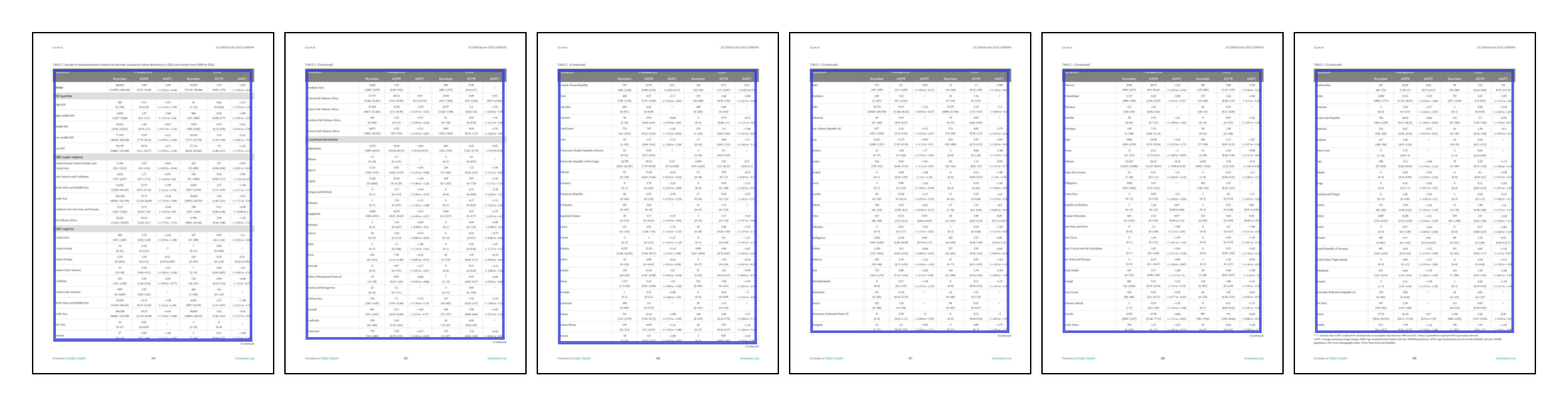} \\
    \includegraphics[height=66px]{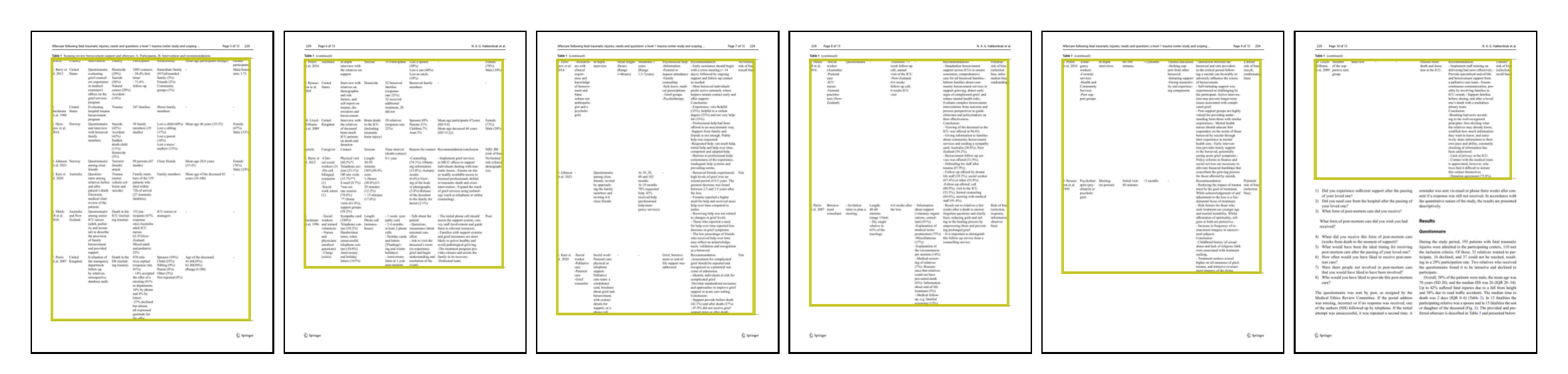} \\
    \includegraphics[height=66px]{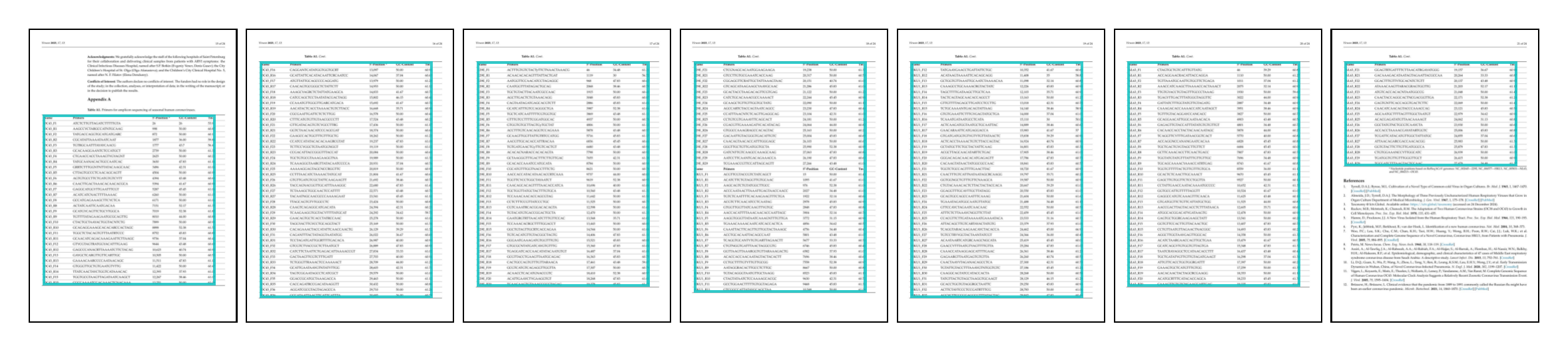} \\
    \caption{This figure shows 5 of the longer multi-page tables from PubTables-v2, with the bounding box for each table highlighted in a particular color on each contiguous page it spans.}
    \label{fig:appendix_multi_page_tables}
\end{figure*}

The third collection of PubTables-v2, the Full Documents collection discussed in \cref{sec:full_documents}, contains 9,172 full documents.
Every document in this collection contains at least one \emph{multi-part} table.
A multi-part table is a table split into two or more parts, either: 1) a multi-page table, split across multiple pages; or 2) a single-page split table, split across multiple columns of a two-column page.

The Full Documents collection contains in total 9,492 multi-page tables, 630 single-page split tables, and 14,740 single-page, single-part tables.
For this collection, we do not annotate bounding boxes for rows or columns within each table, focusing exclusively on table detection and end-to-end table extraction.
Annotating bounding boxes for rows and columns in multi-part tables could have some additional challenges that we leave to future work.

\subsubsection{Annotation}

In this section, we provide additional details about the automated annotation procedure for multi-part tables.
The procedure is identical to the annotation procedure described in \cref{sec:pubtables_v2}, but with one additional step for generating candidate PDF tables.
To locate candidate multi-part tables in PubMed articles, we leverage the published table detection model trained on PubTables-1M \cite{smock2021tabletransformer}.
We run this model on over a million PubMed articles that were not included in PubTables-1M, to detect individual table parts.
To determine whether these parts combine into a multi-part table, we concatenate the text extracted from table parts across contiguous pages, and use the PDF-XML matching procedure detailed in \cref{sec:quality_control} to measure the quality of the match with author-annotated HTML tables for that document.
The criterion for a table match is the same as in \cref{sec:quality_control}.
For including an entire document in this collection, all tables in the XML/HTML annotation must be matched in the PDF and at least one of these must be a multi-part table.

\subsection{Statistics}
\label{sec:appendix_pubtables_v2_statistics}

\begin{table*}[tb]
\caption{\textbf{Multi-page table distribution.} PubTables-v2 contains the first large dataset of multi-page tables annotated for table detection and structure recognition. This table gives a count of tables by the number of pages they span. While most multi-page tables span only 2 pages, over 200 tables span 5 pages or more.}
\centering
\begin{adjustbox}{max width=\linewidth}
\begin{tabular}{l r r r r r r r r r r r r}
\toprule
\textbf{Pages Spanned}  & $\textbf{2}$ & $\textbf{3}$ & $\textbf{4}$ & $\textbf{5}$ & $\textbf{6}$ & $\textbf{7}$ & $\textbf{8}$ & $\textbf{9}$ & $\textbf{10}$ & $\textbf{11}$ & $\textbf{12}$ & $\textbf{13}$ \\
\midrule
\textbf{Tables} & 7,817 & 1,134 & 302 & 109 & 53 & 32 & 23 & 5 & 8 & 6 & 0 & 3 \\
\bottomrule
\end{tabular}
\end{adjustbox}
\label{tab:multi_page_table_distribution}
\end{table*}

\cref{tab:multi_page_table_distribution} gives a breakdown of the total number of multi-page tables by page length in the PubTables-v2 Full Documents collection.
Some of these tables are incredibly long, with 17 tables spanning 10 pages or more.

\begin{figure}[tb]
    \centering
    \includegraphics[width=0.99\linewidth]{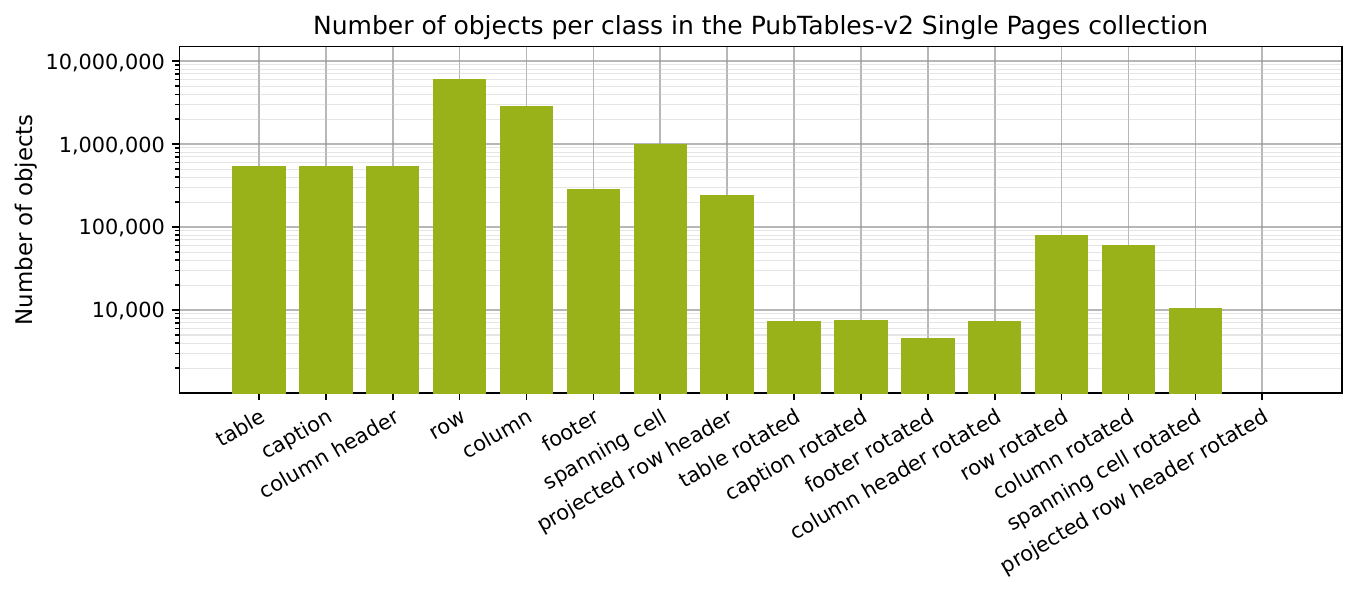}
    \caption{Total number of objects per class in the PubTables-v2 \emph{Single Pages} collection. As can be observed, there is a wide variance in class frequencies. Overall, rotated objects are much rarer than non-rotated objects.}
    \label{fig:objects_per_class}
\end{figure}

\cref{fig:objects_per_class} gives a breakdown of the total number of objects per class in the PubTables-v2 Single Pages collection.
Rotated objects are significantly rarer than non-rotated objects.

\subsection{Annotation Quality Examples}
\label{sec:appendix_annotation_quality}

\begin{figure*}[tb]
    \centering
    \setlength{\fboxsep}{0pt}\setlength{\fboxrule}{0.4pt}
    \begin{tabular}{@{}c@{\hspace{2pt}}c@{\hspace{2pt}}c@{\hspace{2pt}}c@{}}
    \fbox{\includegraphics[height=0.308\textwidth]{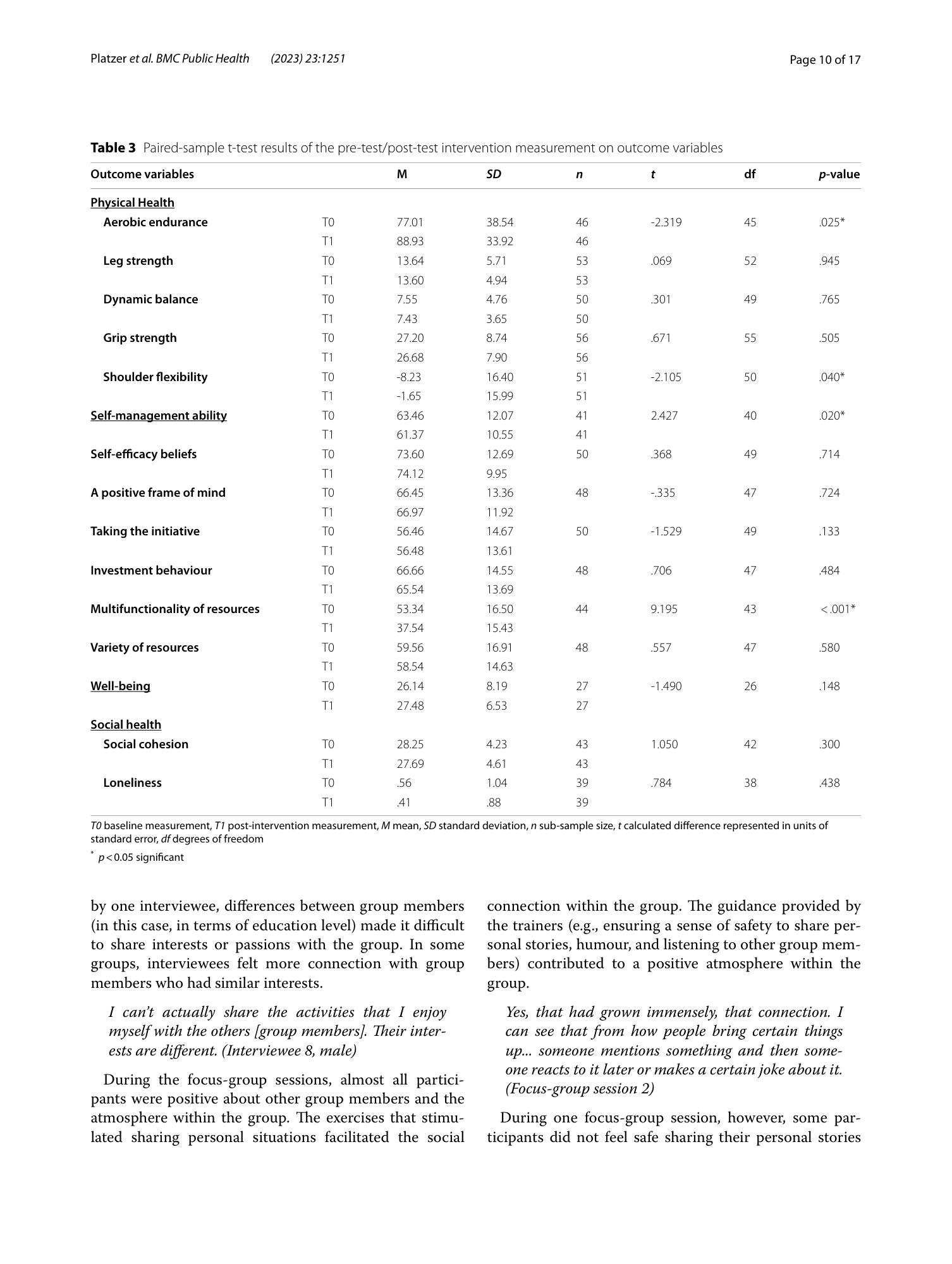}} &
    \fbox{\includegraphics[height=0.308\textwidth]{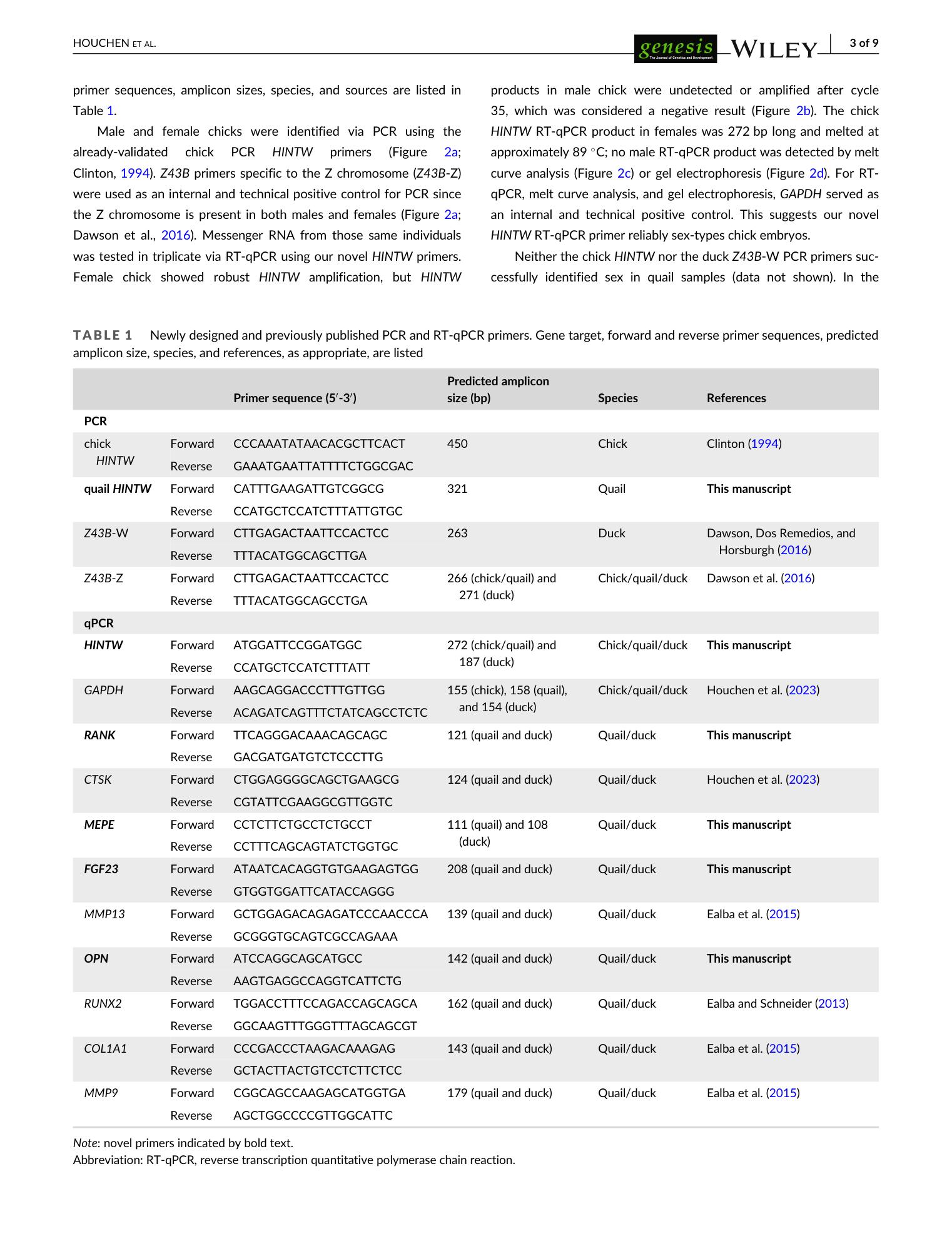}} &
    \fbox{\includegraphics[height=0.308\textwidth]{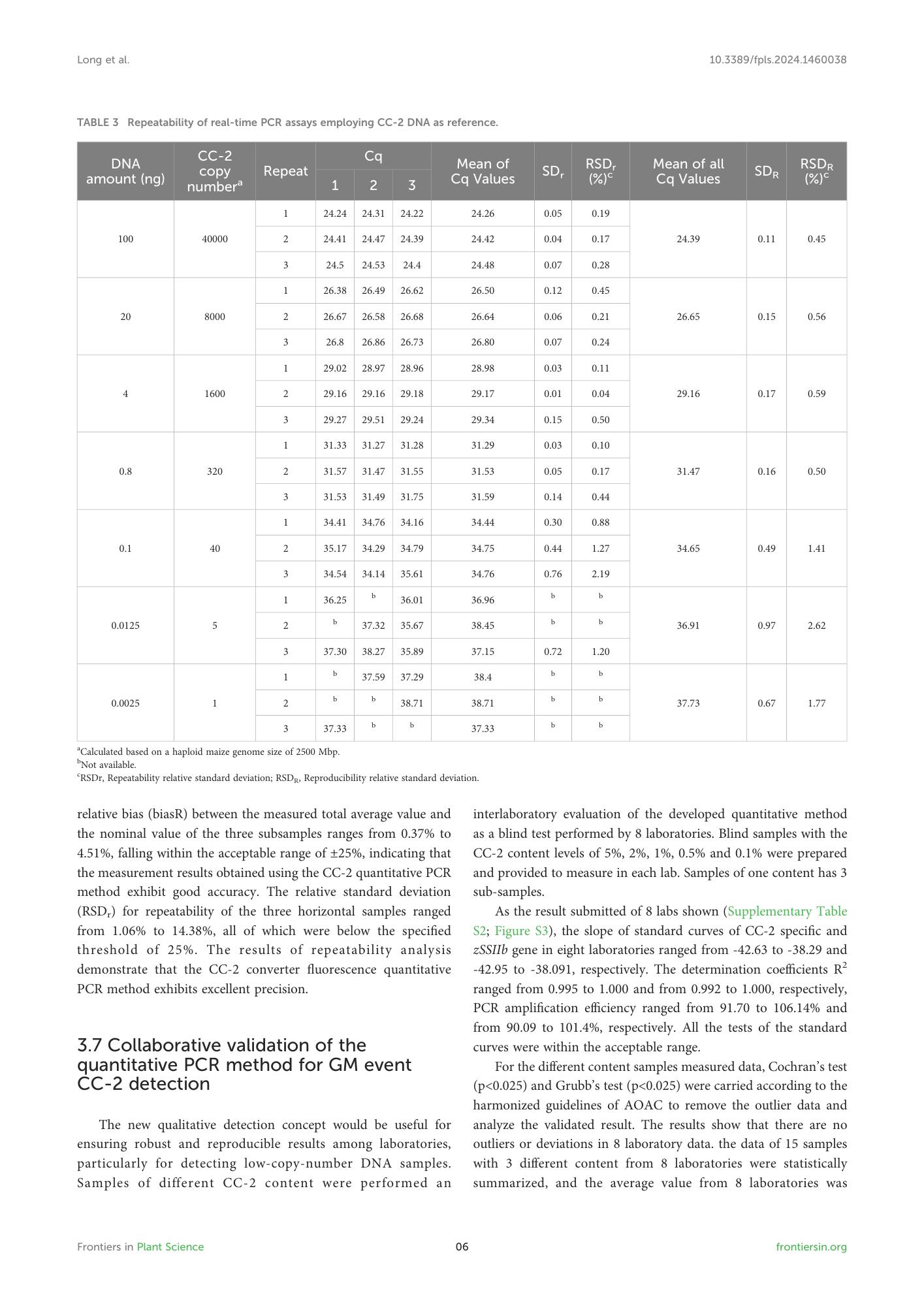}} &
    \fbox{\includegraphics[height=0.308\textwidth]{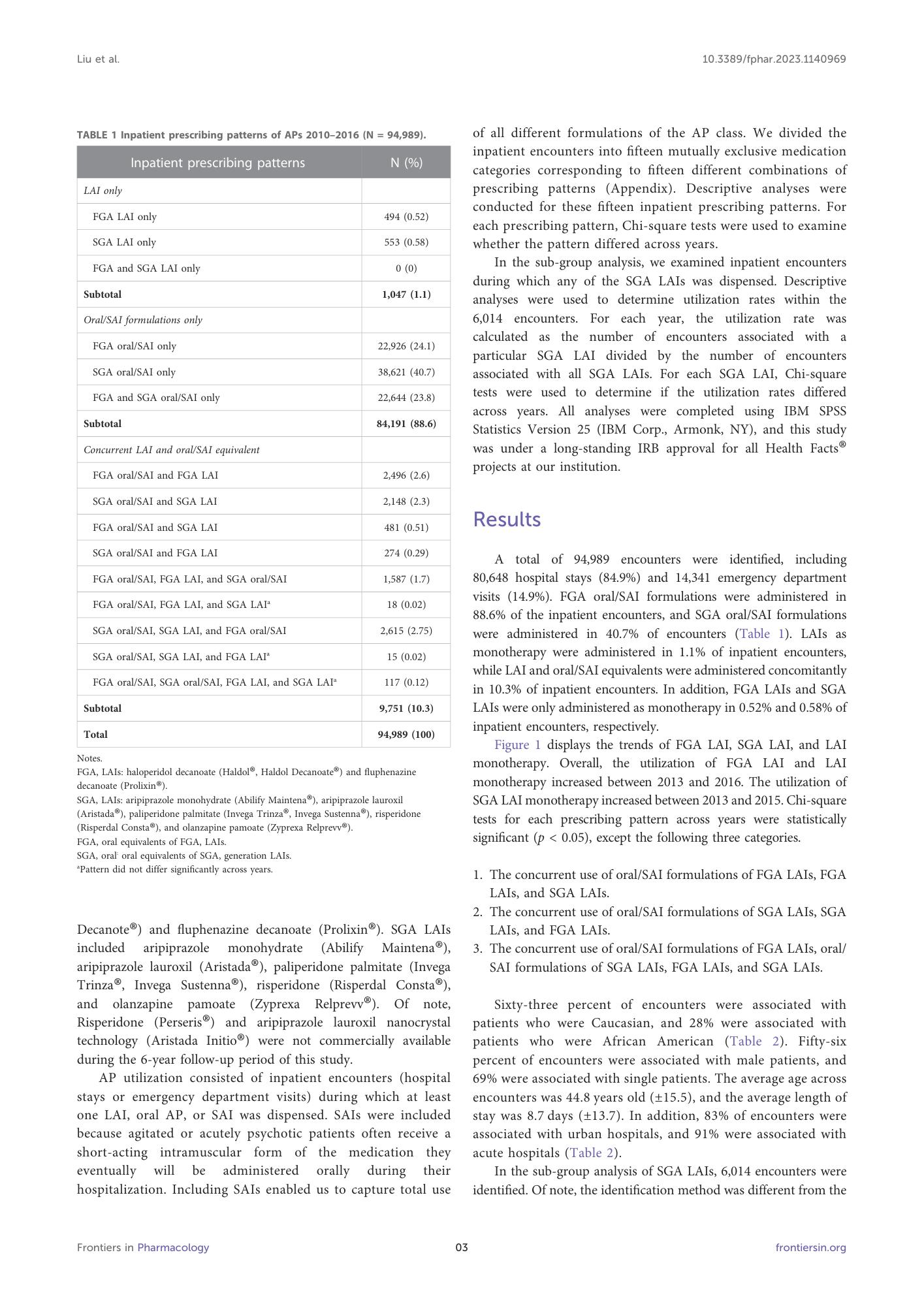}} \\[2pt]
    \fbox{\includegraphics[height=0.308\textwidth]{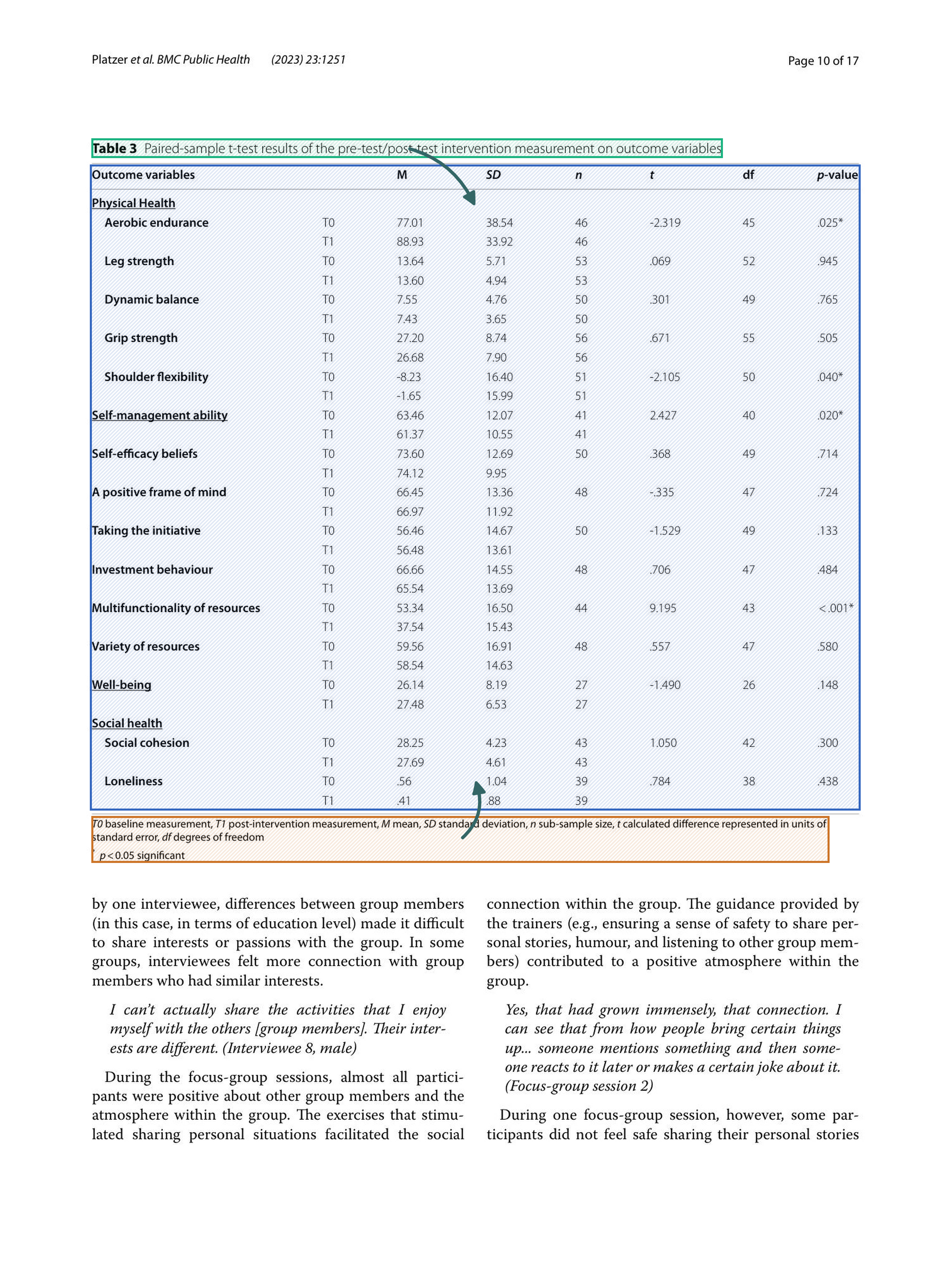}} &
    \fbox{\includegraphics[height=0.308\textwidth]{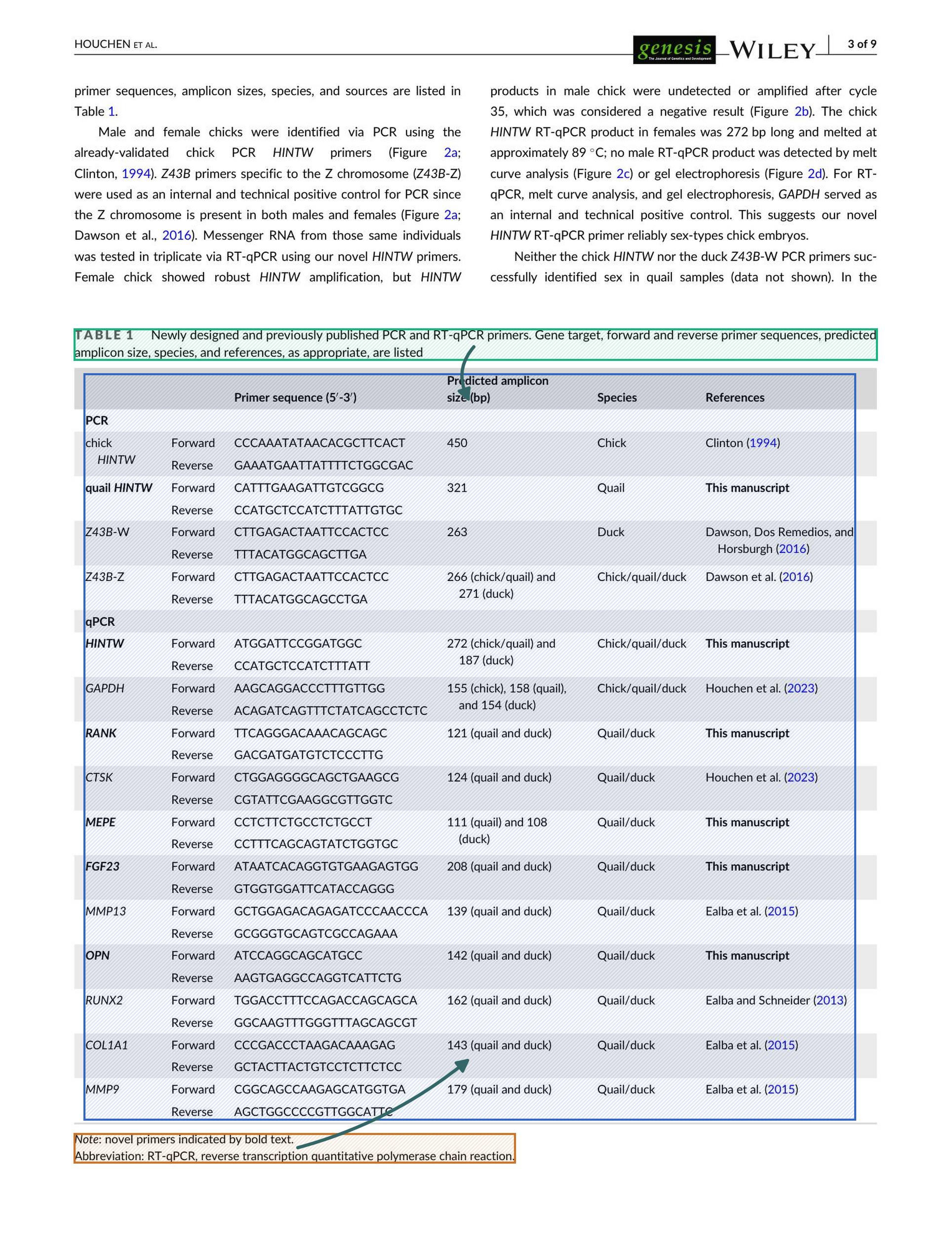}} &
    \fbox{\includegraphics[height=0.308\textwidth]{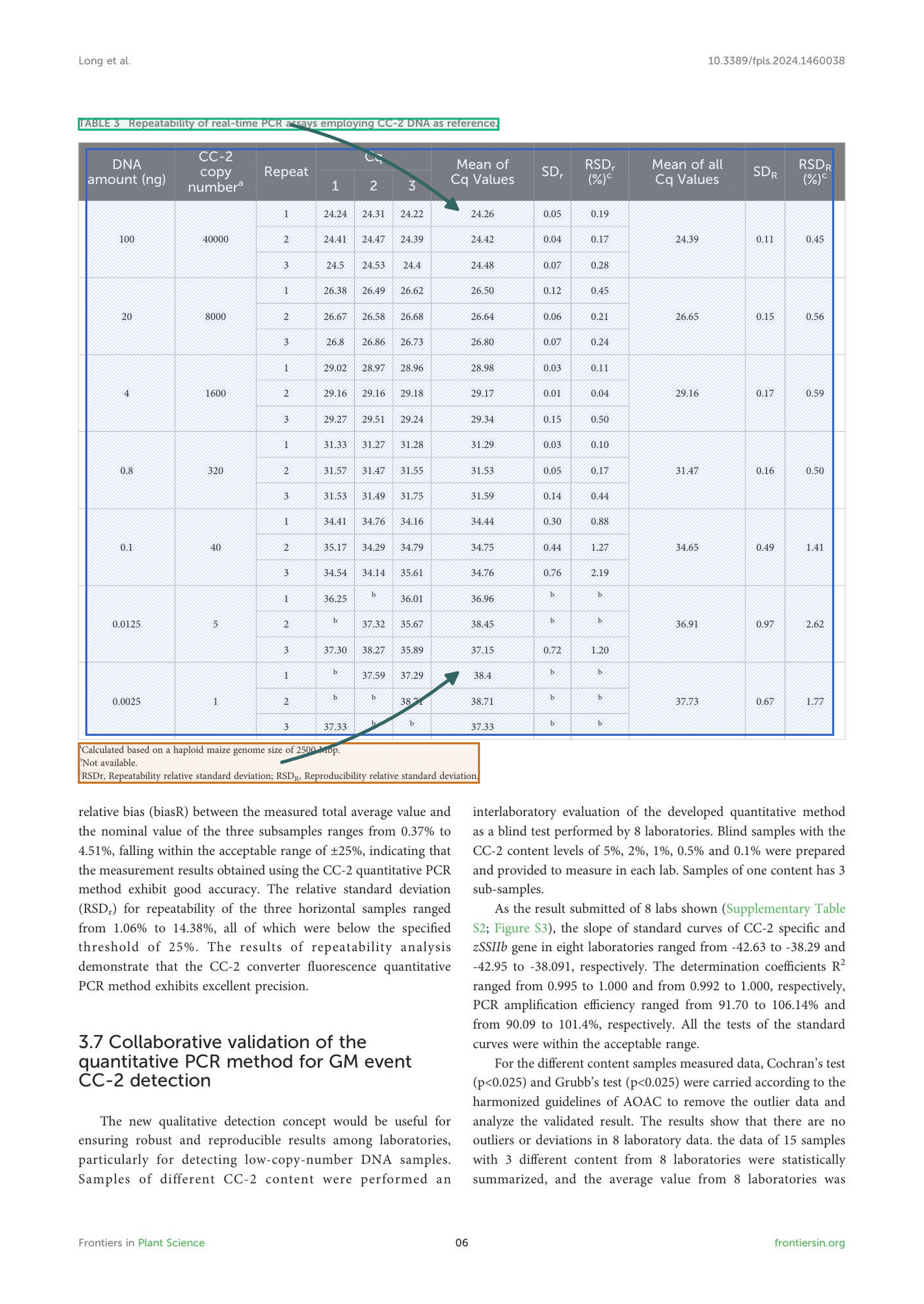}} &
    \fbox{\includegraphics[height=0.308\textwidth]{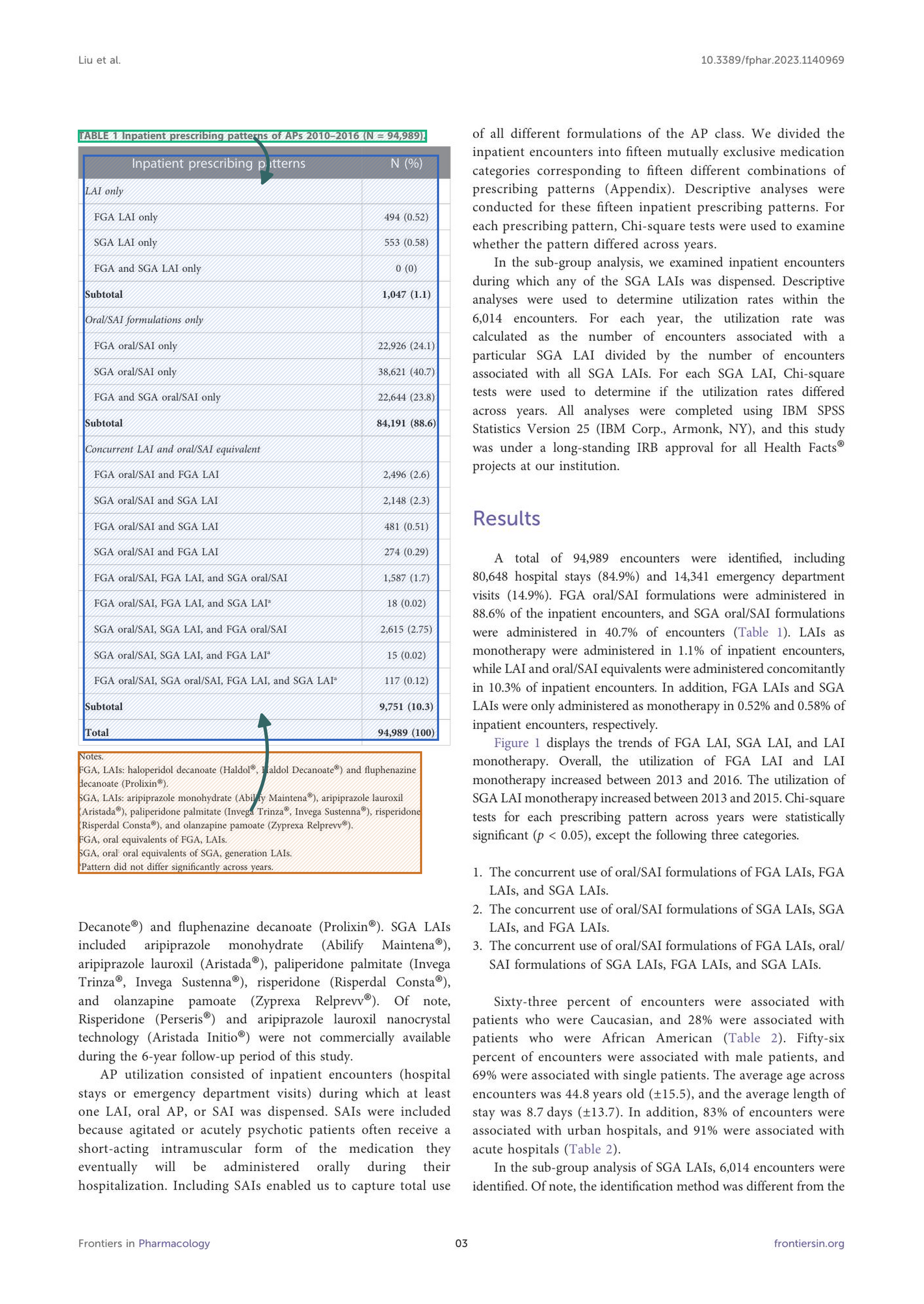}} \\[2pt]
    \fbox{\includegraphics[height=0.308\textwidth]{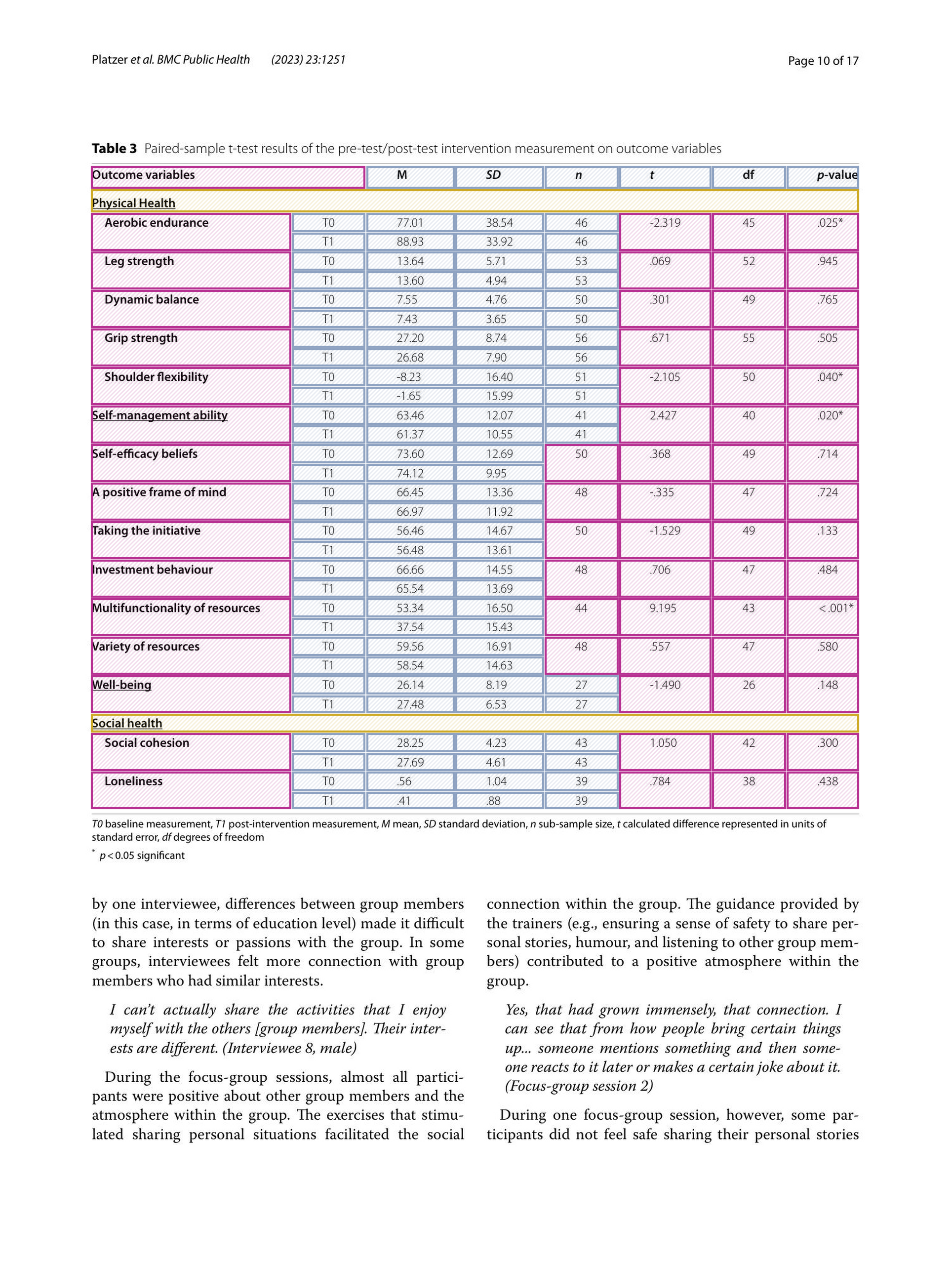}} &
    \fbox{\includegraphics[height=0.308\textwidth]{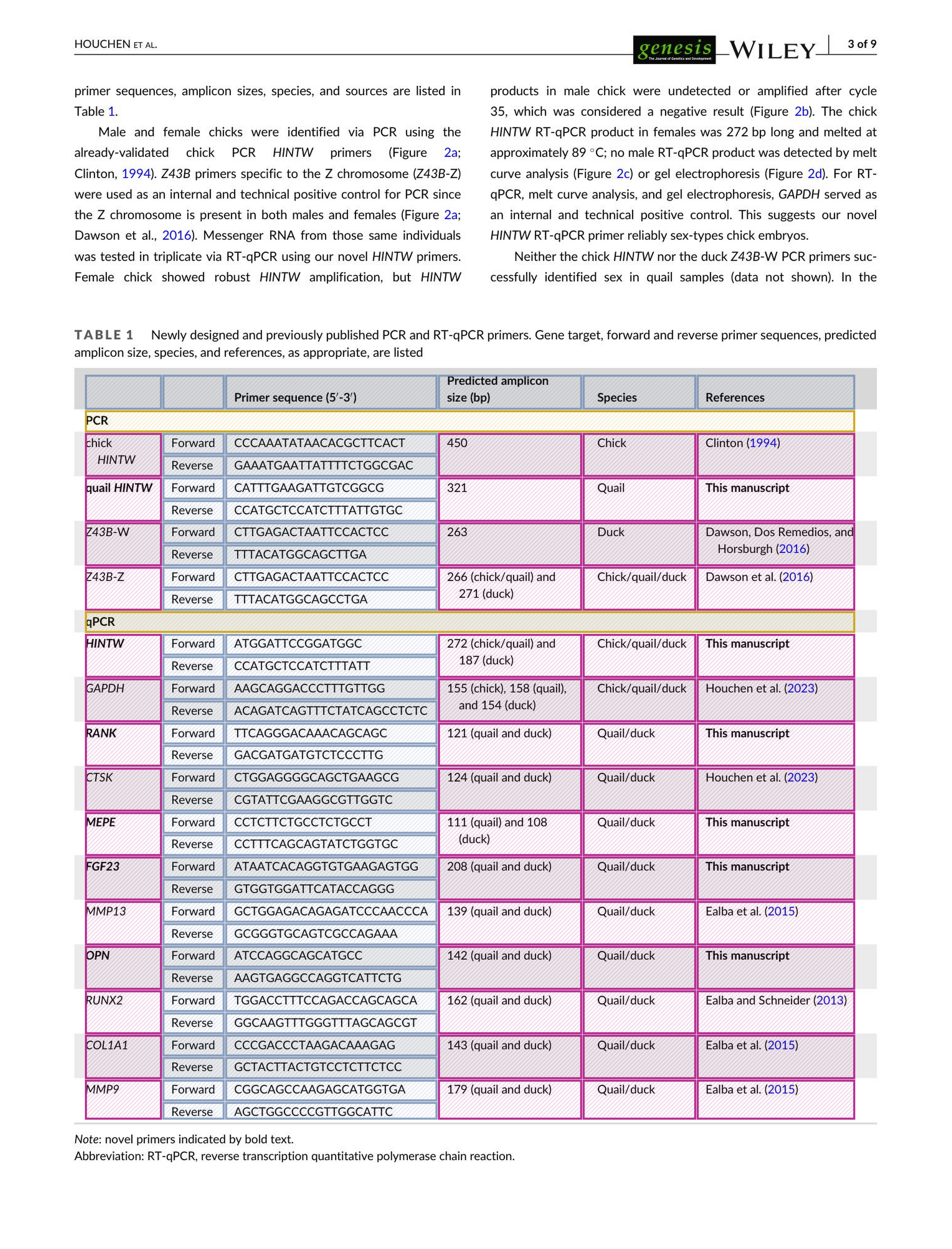}} &
    \fbox{\includegraphics[height=0.308\textwidth]{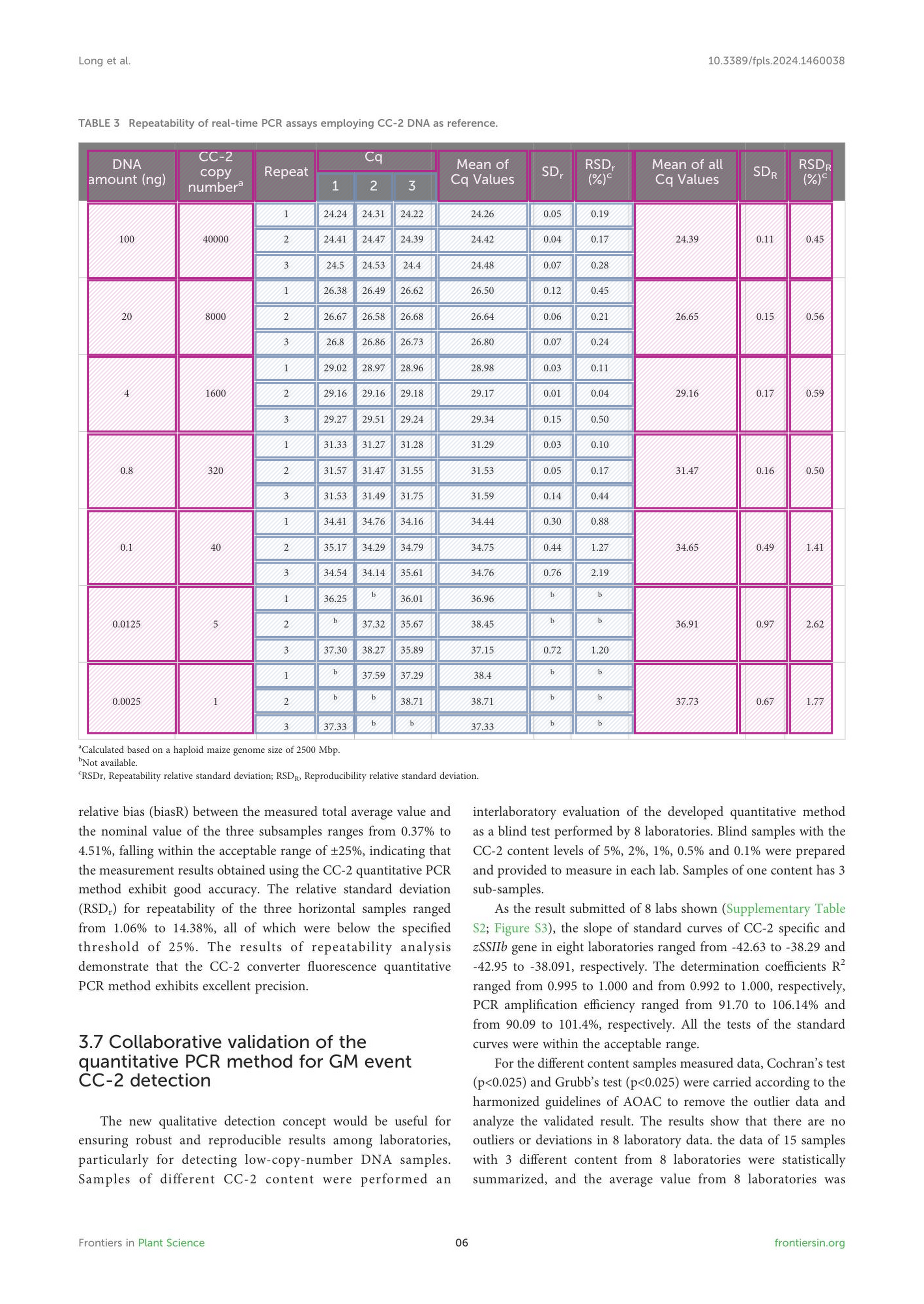}} &
    \fbox{\includegraphics[height=0.308\textwidth]{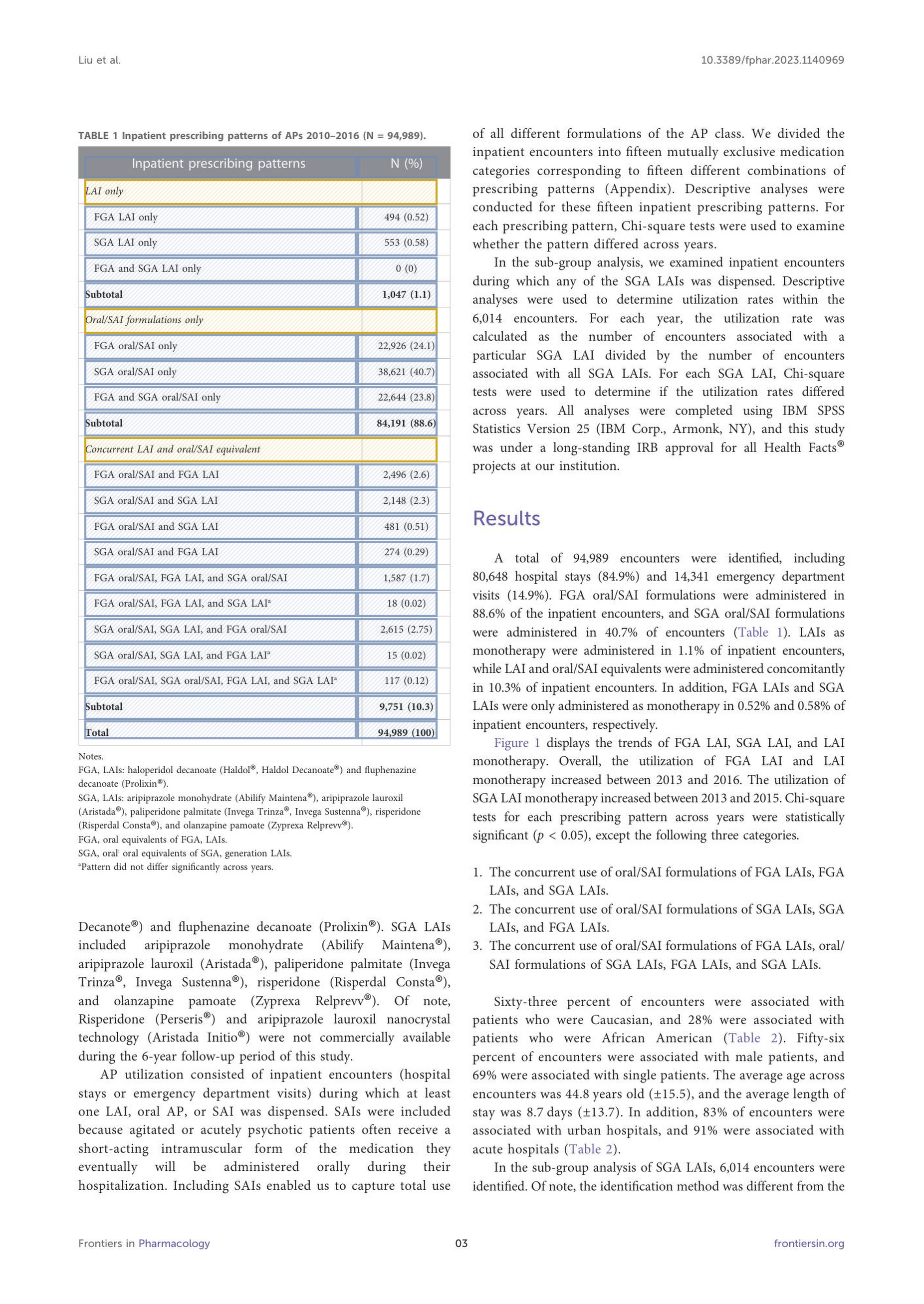}} \\
    \end{tabular}
    \annotlegend
    \caption{\textbf{Annotation layers shown separately} for four pages. Top row: original page images. Middle row: page-level object associations---table (\textcolor{blue}{blue}), caption (\textcolor{teal}{teal}), and footer (\textcolor{orange}{orange}) bounding boxes with arrows indicating which caption/footer belongs to which table. Bottom row: cell-level structure---individual cell bounding boxes colored by type: regular cells (\textcolor{blue!60}{gray-blue}), spanning cells (\textcolor{magenta}{magenta}), and projected row headers (\textcolor{yellow!80!red}{gold}). This decomposition highlights the two complementary annotation layers in PubTables-v2: document-level object relationships and fine-grained table structure.}
    \label{fig:annotation_quality_layers}
\end{figure*}

\cref{fig:annotation_quality_layers} shows the annotation layers in PubTables-v2 for four example pages from the Single Pages collection, illustrating both page-level object associations and cell-level structure annotations.

\section{Metrics}
\label{sec:appendix_metrics}

In this section, we first cover important background details on GriTS \cite{smock2023grits} and TEDS \cite{zhong2020image} metrics, then we describe in detail the page- and document-level versions of these metrics that we use in our experiments.

\subsection{GriTS}
\label{sec:appendix_grits}

GriTS \cite{smock2023grits} is a similarity score between matrices (or, \emph{grids}) for table structure recognition.
It has the general form:

\begin{align}\label{eq:grits}
\text{GriTS}_f(\mathbf{A}, \mathbf{B}) = \frac{2\sum_{i,j} f(\mathbf{\tilde{A}}_{i,j}, \mathbf{\tilde{B}}_{i,j})} {{|\mathbf{A}|} + {|\mathbf{B}|}},
\end{align}

\noindent where $\mathbf{A}$ and $\mathbf{B}$ are the ground truth and predicted matrices (grids), respectively.
Each entry $A_{i,j}$ in each matrix is called a \emph{grid cell}, which represents the intersection of a row and a column in a table.
$f(\mathbf{\tilde{A}}_{i,j}, \mathbf{\tilde{B}}_{i,j})$ is a score function that measures the match between two grid cells based on some property, like its text content, in the range $[0, 1]$.
GriTS has the same form as an F1 score, except that instead of an integer count of true positives, there is a real-valued true positive \emph{score}, given by $\sum_{i,j} f(\mathbf{\tilde{A}}_{i,j}, \mathbf{\tilde{B}}_{i,j})$.
Using the true positive score, pseudo-recall and pseudo-precision can be defined analogously to pseudo-F1.  

\subsection{TEDS}
\label{sec:appendix_teds}
TEDS \cite{zhong2020image} is tree-edit distance similarity. It considers the HTML representation for a table to be a tree, with rows \texttt{<tr>} as the parent node for cells \texttt{<td>}.
Like GriTS, TEDS also measures the match between cell text content, using normalized edit distance.
Unlike GriTS, TEDS takes into account the column header, via column header tags \texttt{<thead>} and \texttt{<th>} in the HTML.
$\textrm{TEDS}_S$, also referred to as TEDS-S or TEDS-Struct, is a version of TEDS that ignores the text content of an HTML tree and considers the cell tree structure only.

\subsection{Extending to Document-Level Table Extraction}
\label{sec:appendix_document_metrics}

GriTS and TEDS are defined for comparing one ground truth table with one predicted table.
This is useful for evaluating TSR on a cropped table where there is only a single ground truth table and a model is assumed to have predicted only one table.
As described in \cref{sec:metrics}, we generalize these metrics from cropped TSR to page- and document-level TE.

We frame document-level TE as a model predicting a \emph{set} of tables, given an input document.
Measuring TE performance then becomes comparing two sets of tables: one set of ground truth tables and one set of predicted tables for that document.
In the most general case, we assume no correspondence is given between the two, nor that they are even given in the same order.
To measure TE quality for a single document, then, we aim to determine the one-to-one match between ground truth and predicted tables that maximizes the total sum of their individual GriTS or TEDS scores, respectively.
This can be done using the standard Hungarian algorithm.

We refer to the sum for either metric as the unnormalized \emph{true positive score}.
In the next section, we describe how we aggregate the unnormalized true positive score for GriTS and TEDS into their final scores.

\subsection{Aggregating GriTS and TEDS Over a Dataset}
\label{sec:appendix_aggregating_grits}

In all cases, whether the input to our model is a cropped table, a page, or a document, there are zero or more ground truth tables and the model predicts a set of zero or more tables.
We apply the Hungarian algorithm to maximize GriTS or TEDS for a single input sample, yielding an unnormalized true positive score for that sample.
For a collection of samples, we sum the true positive score over all samples, and compute GriTS and TEDS as a pseudo-F1 score.
In the case of TEDS, recall is the total true positive score divided by the total number of ground truth tables over all samples.
Similarly, precision is the total true positive score divided by the total number of predicted tables, over all input samples.

For GriTS, we compute all three similarly, but with an additional weighting per table: we weight the true positive score, as well as the total true positives and total predictions for each table, by the number of cells in the table.
This is straightforward to do for GriTS, as the grid matrices always make the number of cells unambiguous.
For TEDS, the number of cells may not be unambiguous for all possible parseable HTML sequences and valid trees.

Note that this way of aggregating GriTS over a collection of samples is not the same as the original definition by Smock \etal~\cite{smock2023grits}.
In the original definition, GriTS was a pseudo-F1 score for a single input (more specifically, a single cropped table), but the aggregated score was the mean GriTS score over all inputs.
The new definition aggregates the true positive score over all samples, then computes the pseudo-F1 score.

\subsection{Graph Metrics}
\label{sec:appendix_graph_metrics}

To evaluate graph predictions, we use the F1 metric for edges defined by DocParser \cite{rausch2021docparser}.
Under this metric, a predicted edge (or relation triple) counts as a true positive if its source and target nodes match the source and target nodes of a ground truth edge exactly.
For an exact match, the class labels for each corresponding node must be identical, and the intersection-over-union (IoU) for the bounding boxes of the nodes must be above a certain threshold.
\cite{rausch2021docparser} evaluate this metric for IoU thresholds of 0.5, 0.65, and 0.8.
In this work, we evaluate using just the highest IoU threshold of 0.8.

\section{Models}
\label{sec:appendix_models}

In this section, we provide additional details about the models used in our experiments.

\subsection{TATR-v1.2-Pub}
\label{sec:appendix_tatr}

For this model, we adopt all the same architecture parameters and initialize the model weights from \texttt{TATR-v1.1-Pub} \cite{smock2023aligning}.
We then fine-tune it on the PubTables-v2 Cropped Tables subset.
For training details, see \cref{sec:appendix_training} and \cref{sec:appendix_tatr_training}.

\subsection{VLMs}
\label{sec:appendix_vlms}

\subsubsection{SmolDocling and GraniteDocling}

We use \texttt{SmolDocling} \cite{nassar2025smoldocling} (\texttt{\seqsplit{docling-project/SmolDocling-256M-preview}} on Hugging Face) and its successor \texttt{\seqsplit{GraniteDocling}} (\texttt{\seqsplit{ibm-granite/granite-docling-258M}} on Hugging Face).
For inference, we use vLLM, which significantly speeds up inference.
For document understanding tasks, both models were trained to produce a custom DocTags output format.
We use the \texttt{docling-core} Python package to parse the DocTags model output and convert it into HTML format.
\cref{tab:doctags_parsing_errors} shows that the model output is not always parsable.
The Single Pages collection seems to be especially challenging for the SmolDocling model.
Its successor, GraniteDocling, handles the Single Pages collection substantially better.
However, GraniteDocling produces marginally more errors on the Cropped Tables collection.
We use the \texttt{beautifulsoup4} Python package to parse the HTML output produced by \texttt{docling-core}.

\begin{table*}[tb]
\caption{
\textbf{DocTags Parsing Errors.} This table shows the total numbers of parsing errors for
 DocTags outputs of \texttt{SmolDocling} and \texttt{GraniteDocling}.
We use the \texttt{docling-core} Python package (version \texttt{2.48.4}) to parse the DocTags model outputs.
}
\centering
\begin{tabular}{l r r r}
\toprule
\multirow{2}{*}{\textbf{Collection}} & \multicolumn{2}{c}{\textbf{No. of Parsing Errors}} & \multirow{2}{*}{\textbf{No. of Test Samples}} \\
\noalign{\vspace{0.5ex}}
\cline{2-3}
\noalign{\vspace{0.6ex}}
               & \textbf{SmolDocling} & \textbf{GraniteDocling} & \\
\midrule
Cropped Tables & 16 & 43 &  7,680 \\
Single Pages   & 958 & 35 & 7,680\\
\bottomrule
\end{tabular}
\label{tab:doctags_parsing_errors}
\end{table*}

\subsubsection{Qwen2.5-VL-3B}

We use \texttt{Qwen2.5-VL-3B} \cite{bai2025qwen2} (\emph{Qwen/Qwen2.5-VL-3B-Instruct} on Hugging Face).
We use vLLM to speed up inference.
For document parsing, Qwen-VL models are capable of producing outputs in a QwenVL HTML format, which augments the standard HTML format with additional information about the spatial layout of the document (e.g. coordinates of the table on the page).
We use the \texttt{beautifulsoup4} Python package to parse the model output.
Since \texttt{beautifulsoup4} is designed to parse even malformed or truncated HTML, parsing never fails even if the model generates an invalid HTML document.
We evaluate \texttt{Qwen2.5-VL-3B} on all three collections (Cropped Tables, Single Pages, and Full Documents).
For the Full Documents collection, we add images of all document pages to the VLM input.
When inputting all the pages of a document at once, \texttt{Qwen2.5-VL-3B} is the only VLM out of the models we evaluated that is able to process nearly all documents from the Full Documents collection's test set.
In our experiments, of the test set's 878 documents, \texttt{Qwen2.5-VL-3B} fails to process only the two longest documents (65 and 72 pages long).

\subsubsection{Granite-Vision-3.2-2B}

We use \texttt{Granite Vision} \cite{team2025granite} (\emph{ibm-granite/granite-vision-3.2-2b} on Hugging Face). We use the off-the-shelf model for inference, prompting the model to produce markdown tables. We parsed the markdown tables, including their cell span information when present, for evaluation. The Granite Vision model produces its own format of spanning information for cells that span multiple columns or rows, and we parse this information. The format of this generated output resembles the following:

\begin{lstlisting}[
    basicstyle=\ttfamily\small
]
<md> | <ROWSPAN=2> Cell | <COLSPAN=2> Cell |
     | Cell | Cell | Cell |
     | --- | --- | --- |
     | Data | Data | Data | </md>
\end{lstlisting}

We test a variety of prompts when running inference, either focusing solely on table extraction or full page extraction, and testing both markdown and HTML. We find a table-specific markdown prompt to be the most reliable.

\section{Training}
\label{sec:appendix_training}

\texttt{TATR-v1.2-Pub} is trained on 8 Nvidia T4 GPUs with a batch size of 2 on each GPU, for an effective batch size of 16.
The model weights are initialized from \texttt{TATR-v1.1-Pub} \cite{smock2023aligning}, a pre-trained model trained on cropped tables from PubTables-1M.

\subsection{TATR-v1.2-Pub}
\label{sec:appendix_tatr_training}

For \texttt{TATR-v1.2-Pub}, we consider an epoch to be 100,000 samples from the PubTables-v2 Cropped Tables collection.
Under this definition, the model is trained for 160 epochs, with an initial learning rate of 0.00005 (5e-5) and a learning rate gamma of 0.9 applied every 4 epochs.
We generally use the default hyperparameters from TATR \cite{smock2022pubtables}, except the weight on the "no object" class, \texttt{eos\_coef}, which we set to 0.3.

\section{Additional Experiments and Results}
\label{sec:pubtables_v2_appendix}

\subsection{Extended Qualitative Analysis}
\label{sec:appendix_qualitative_gallery}

\begin{figure*}[tb]
    \centering
    \setlength{\fboxsep}{0pt}\setlength{\fboxrule}{0.4pt}
    \fbox{\includegraphics[width=0.47\textwidth]{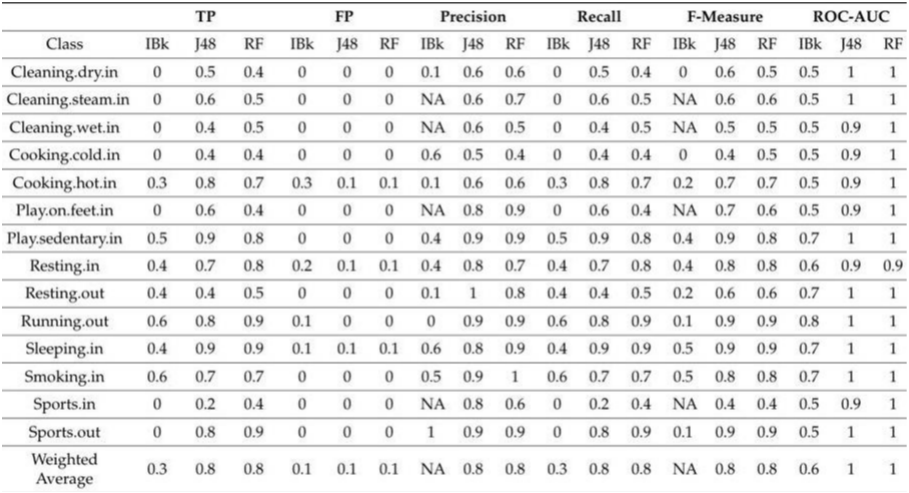}} \\
    {\footnotesize (a) Original table} \\[4pt]
    \begin{tabular}{@{}c@{\hspace{4pt}}c@{}}
    \fbox{\includegraphics[width=0.47\textwidth]{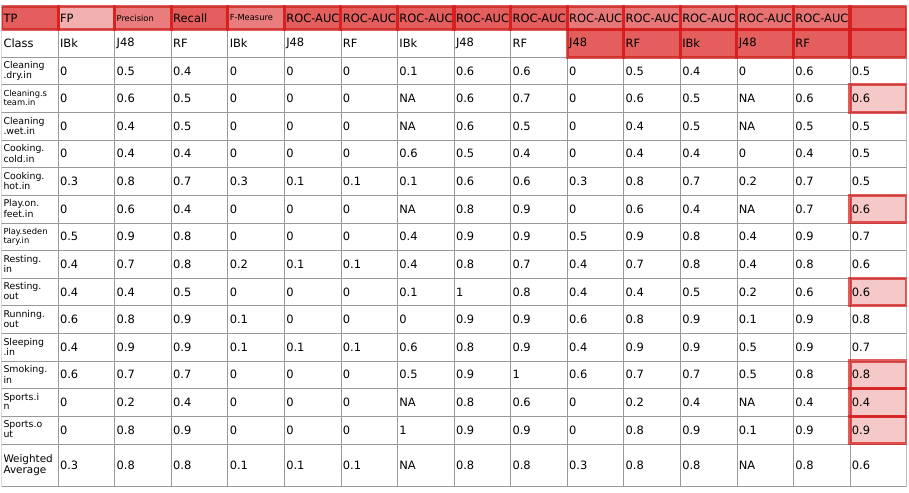}} &
    \fbox{\includegraphics[width=0.47\textwidth]{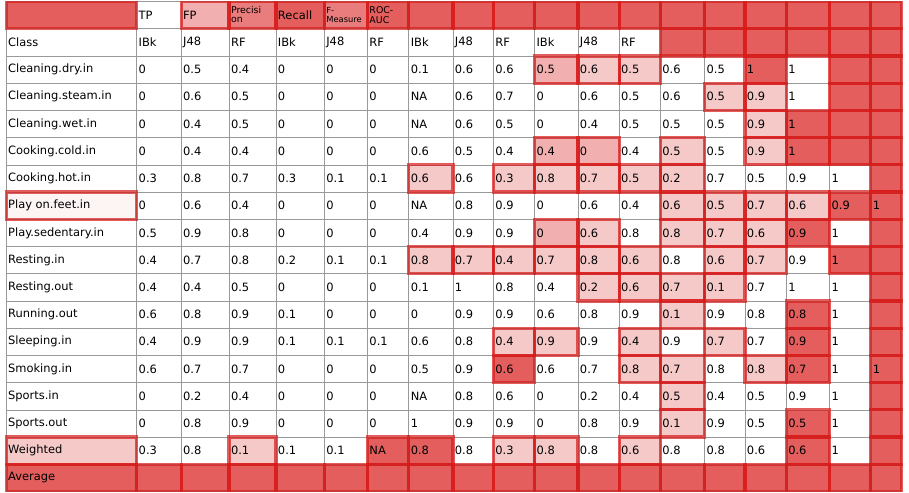}} \\
    {\footnotesize (b) SmolDocling} &
    {\footnotesize (c) Qwen2.5-VL} \\[4pt]
    \fbox{\includegraphics[width=0.47\textwidth]{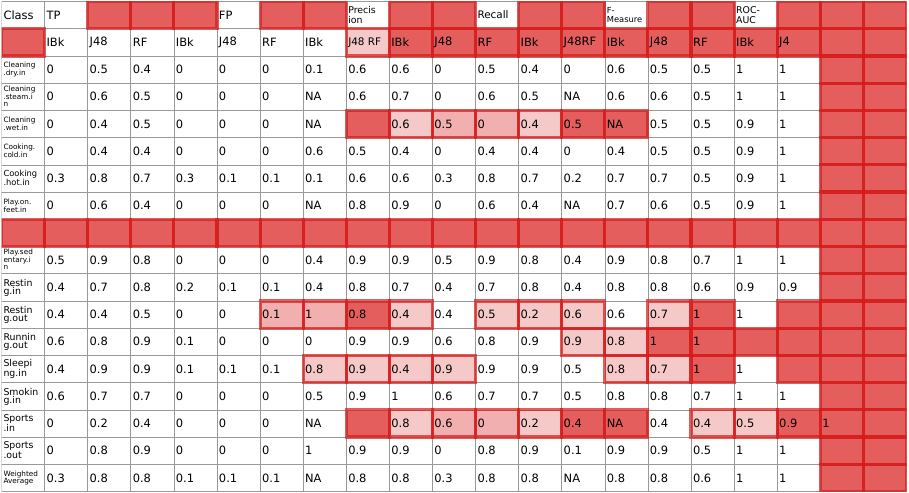}} &
    \fbox{\includegraphics[width=0.47\textwidth]{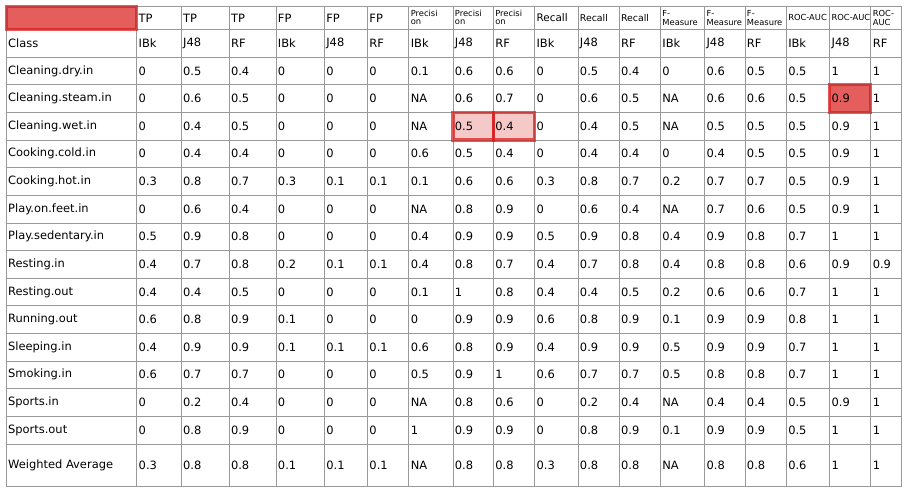}} \\
    {\footnotesize (d) DeepSeek-OCR 2} &
    {\footnotesize (e) Claude Opus 4.6} \\
    \end{tabular}
    \par\vspace{4pt}
    \precisioncolorbar
    \caption{\textbf{Qualitative failure analysis} for table PMC10747712\_table\_3 from the Cropped Tables collection (17 rows $\times$ 19 columns). Cell-level GriTS precision grids are shown for four models, with red intensity indicating content error magnitude. Errors exhibited include numeric quantity failures, additional rows, additional columns, cell misalignments, and spanning cell failures.
    Note that SmolDocling (b) drops some columns entirely, which contributes to a lower score but cannot be visualized here.}
    \label{fig:qualitative_gallery_1}
\end{figure*}

\cref{fig:qualitative_gallery_1} provides an additional qualitative example complementing the failure analysis in the main paper (\cref{fig:precision_qualitative_1row}). This example from the Cropped Tables collection demonstrates diverse failure modes on a wide table: structural errors (missing or extra rows/columns) versus content-only errors, with performance varying widely across model families.

\subsection{Complete Evaluation Results}
\label{sec:appendix_full_results}

The main paper reports abbreviated results tables for space.
Here we present the complete evaluation results including all metrics and all models evaluated: cropped table TSR (\cref{tab:appendix_experiment_tsr}), page-level TE (\cref{tab:appendix_page_level_results}), and document-level TE (\cref{tab:appendix_document_level_results}).

\begin{table*}[tb]
\caption{\textbf{Cropped table structure recognition (complete results).} Full evaluation results on the PubTables-v2 Cropped Tables collection containing long and wide tables, including all metrics and models.}
\centering
\begin{adjustbox}{max width=\linewidth}
{\footnotesize
\begin{tabular}{l r r r r r r}
\toprule
\textbf{Model} & $\textbf{GriTS}_{\textbf{Top}}$ & $\textbf{GriTS}_{\textbf{Con}}$ & $\textbf{Acc}_{\textbf{Top}}$ & $\textbf{Acc}_{\textbf{Con}}$ & $\textbf{TEDS}_{\textbf{S}}$ & $\textbf{TEDS}$ \\
\midrule
SmolDocling-256M \cite{nassar2025smoldocling} & 0.2523 & 0.1923 & 0.1996 & 0.0250 & 0.7102 & 0.5928 \\
GraniteDocling-258M \cite{granitedocling2025} & 0.5482 & 0.5026 & 0.2065 & 0.0555 & 0.6699 & 0.6284 \\
Qwen2.5-VL-3b \cite{bai2025qwen2} & 0.6586 & 0.5305 & 0.1751 & 0.0064 & 0.7331 & 0.6672 \\
Granite-Vision-3.2-2b \cite{team2025granite} & 0.8714 & 0.7542 & 0.2155 & 0.0067 & 0.7769 & 0.6953 \\
DeepSeek-OCR \cite{wei2025deepseek} & 0.8115 & 0.7541 & 0.2400 & 0.0264 & 0.7453 & 0.7109 \\
DeepSeek-OCR 2 \cite{wei2026deepseek} & 0.7953 & 0.7429 & 0.2148 & 0.0294 & 0.7575 & 0.7148 \\
dots.ocr \cite{li2025dots} & 0.8343 & 0.8011 & 0.3190 & 0.0566 & 0.7994 & 0.7619 \\
\midrule
Claude Opus 4.6 & 0.9058 & 0.8952 & 0.3389 & 0.1060 & 0.8710 & 0.8512 \\
GPT-5.4 & 0.8842 & 0.8775 & 0.3124 & 0.0988 & 0.8397 & 0.8215 \\
Gemini 3.1 Pro & 0.9354 & 0.9242 & 0.4518 & 0.1613 & 0.9125 & 0.8964 \\
\midrule
TATR-v1.1-Fin \cite{smock2023aligning} + DT & 0.8985 & 0.8683 & 0.1405 & 0.1190 & 0.8684 & 0.8237 \\
TATR-v1.1-Pub \cite{smock2023aligning} + DT & 0.9608 & 0.9585 & 0.4867 & 0.4816 & 0.9481 & 0.9396 \\
\midrule
\multicolumn{7}{l}{\textbf{Trained on PubTables-v2}} \\
\quad TATR-v1.2-Pub + PaddleOCR \cite{cui2025paddleocr} & 0.9801 & 0.9007 & 0.6835 & 0.0044 & 0.9710 & 0.8809 \\
\quad TATR-v1.2-Pub + docTR \cite{doctr2021} & 0.9802 & 0.9286 & 0.6875 & 0.0189 & 0.9713 & 0.9231 \\
\quad TATR-v1.2-Pub + DT & 0.9806 & 0.9801 & 0.6892 & 0.6857 & 0.9717 & 0.9690 \\
\bottomrule
\end{tabular}}
\end{adjustbox}
\label{tab:appendix_experiment_tsr}
\end{table*}

\begin{table*}[tb]
\caption{\textbf{Page-level table extraction (complete results).} Full evaluation results on the PubTables-v2 Single Pages collection, including all metrics and models.}
\centering
\begin{adjustbox}{max width=\linewidth}
\begin{tabular}{l r r r r r r}
\toprule
\textbf{Model} & $\textbf{GriTS}_{\textbf{Top}}$ & $\textbf{GriTS}_{\textbf{Con}}$ & $\textbf{Acc}_{\textbf{Top}}$ & $\textbf{Acc}_{\textbf{Con}}$ & $\textbf{TEDS}_{\textbf{S}}$ & $\textbf{TEDS}$ \\
\midrule
SmolDocling-256M & 0.2502 & 0.2177 & 0.2800 & 0.0298 & 0.5980 & 0.5250 \\
GraniteDocling-258M & 0.7252 & 0.6670 & 0.3794 & 0.0829 & 0.6685 & 0.6155 \\
Qwen2.5-VL-3b & 0.4672 & 0.3803 & 0.3651 & 0.0619 & 0.2859 & 0.2632 \\
Granite-Vision-3.2-2b & 0.8015 & 0.7480 & 0.4245 & 0.0331 & 0.8665 & 0.8092 \\
DeepSeek-OCR & 0.8629 & 0.8207 & 0.6184 & 0.1671 & 0.7996 & 0.7639 \\
DeepSeek-OCR 2 & 0.8839 & 0.8386 & 0.5258 & 0.1120 & 0.7688 & 0.7263 \\
dots.ocr & 0.9241 & 0.8991 & 0.6425 & 0.1872 & 0.9168 & 0.8743 \\
\midrule
Claude Opus 4.6 & 0.9146 & 0.9106 & 0.4934 & 0.2713 & 0.9073 & 0.8945 \\
GPT-5.4 & 0.9208 & 0.9102 & 0.5461 & 0.2074 & 0.9084 & 0.8874 \\
Gemini 3.1 Pro & 0.9500 & 0.9418 & 0.6348 & 0.3524 & 0.9392 & 0.9250 \\
\midrule
\multicolumn{7}{l}{\textbf{Trained on PubTables-v2}} \\
\quad POTATR + PaddleOCR & 0.9614 & 0.8768 & 0.6953 & 0.0060 & 0.9545 & 0.8780 \\
\quad POTATR + docTR & 0.9606 & 0.8809 & 0.6724 & 0.0684 & 0.9511 & 0.8971 \\
\quad POTATR + DT & 0.9674 & 0.9647 & 0.7012 & 0.6447 & 0.9576 & 0.9516 \\
\bottomrule
\end{tabular}
\end{adjustbox}
\label{tab:appendix_page_level_results}
\end{table*}

\begin{table*}[tb]
\caption{\textbf{Document-level table extraction (complete results).} Full evaluation results on the PubTables-v2 Full Documents collection, including all metrics and models.}
\centering
\begin{adjustbox}{max width=\linewidth}
\begin{tabular}{l r r r r r r}
\toprule
\textbf{Model/Context} & $\textbf{GriTS}_{\textbf{Top}}$ & $\textbf{GriTS}_{\textbf{Con}}$ & $\textbf{Acc}_{\textbf{Top}}$ & $\textbf{Acc}_{\textbf{Con}}$ & $\textbf{TEDS}_{\textbf{S}}$ & $\textbf{TEDS}$ \\
\midrule
\multicolumn{7}{l}{\textbf{Full document}} \\
\quad Claude Opus 4.6 & 0.9116 & 0.9079 & 0.5799 & 0.2452 & 0.8914 & 0.8798 \\
\quad GPT-5.4 & 0.8865 & 0.8847 & 0.5038 & 0.1636 & 0.8563 & 0.8341 \\
\quad Gemini 3.1 Pro & 0.9365 & 0.9309 & 0.4854 & 0.2025 & 0.9048 & 0.8905 \\
\multicolumn{7}{l}{\textbf{Pages processed individually}} \\
\quad SmolDocling-256M & 0.1304 & 0.1107 & 0.2471 & 0.0335 & 0.5112 & 0.4110 \\
\quad GraniteDocling-258M & 0.5260 & 0.4723 & 0.2732 & 0.0728 & 0.5173 & 0.4470 \\
\quad Qwen2.5-VL-3b & 0.4215 & 0.2962 & 0.2835 & 0.0452 & 0.1860 & 0.1622 \\
\quad Granite-Vision-3.2-2b & 0.4257 & 0.3382 & 0.2757 & 0.0218 & 0.2658 & 0.2246 \\
\quad DeepSeek-OCR & 0.4548 & 0.4068 & 0.3649 & 0.0921 & 0.5565 & 0.5091 \\
\quad DeepSeek-OCR 2 & 0.4920 & 0.4527 & 0.3661 & 0.0933 & 0.5628 & 0.5176 \\
\quad dots.ocr & 0.6054 & 0.5768 & 0.3958 & 0.1180 & 0.6179 & 0.5876 \\
\multicolumn{7}{l}{\textbf{Pages with cross-page merging (\cref{sec:exp_cross_page_tables})}} \\
\quad dots.ocr + merging & 0.7772 & 0.7495 & 0.5059 & 0.1180 & 0.8022 & 0.7731 \\
\bottomrule
\end{tabular}
\end{adjustbox}
\label{tab:appendix_document_level_results}
\end{table*}

\subsection{Single-Page Split-Table Extraction}
\label{sec:appendix_split_table_extraction}

\begin{table*}[tb]
\caption{\textbf{Single-page split table.} Performance of models on tables split into multiple parts within a single page. There are 64 examples of these in the test set, with each table extending across both columns of a two-column page.}
\centering
\begin{adjustbox}{max width=\linewidth}
\begin{tabular}{l r r r r r r}
\toprule
\textbf{Model} & $\textbf{GriTS}_{\textbf{Top}}$ & $\textbf{GriTS}_{\textbf{Con}}$ & $\textbf{Acc}_{\textbf{Top}}$ & $\textbf{Acc}_{\textbf{Con}}$ & $\textbf{TEDS}_{\textbf{S}}$ & $\textbf{TEDS}$ \\
\midrule
SmolDocling-256M & 0.5583 & 0.4766 & 0.0469 & 0.0000 & 0.4377 & 0.2711 \\
GraniteDocling-258M & 0.5765 & 0.5174 & 0.0469 & 0.0000 & 0.2057 & 0.1705 \\
Qwen2.5-VL-3b & 0.4627 & 0.4230 & 0.0313 & 0.0000 & 0.2108 & 0.1615 \\
Granite-Vision-3.2-2b & \textbf{0.6961} & \textbf{0.6323} & 0.0313 & 0.0000 & \textbf{0.6075} & \textbf{0.5160} \\
DeepSeek-OCR & 0.5230 & 0.4923 & 0.0313 & 0.0000 & 0.3978 & 0.3471 \\
DeepSeek-OCR 2 & 0.6278 & 0.5859 & 0.0313 & 0.0000 & 0.4182 & 0.3667 \\
dots.ocr & 0.6668 & 0.6171 & \textbf{0.0625} & 0.0000 & 0.4427 & 0.3946 \\
\bottomrule
\end{tabular}
\end{adjustbox}
\label{tab:single_page_split_table_results}
\end{table*}

In this experiment, we evaluate the models run page-by-page on the Full Documents collection in \cref{sec:doc_level_te}, but only evaluating pages that contain a single-page split table, like the one on page 4 in \cref{fig:full_document}.
Since models in the previous experiment only see one page at a time, which prevents fully recognizing multi-page tables, this experiment tests whether models can handle tables split into multiple parts within a single page.
The results are given in \cref{tab:single_page_split_table_results}.

The model that appears to perform the best according to most metrics is \texttt{Granite-Vision-3.2-2b}.
However, one caveat is that we noticed in our experiments that this model always produced only a single table as output, no matter the input.
Thus in this experiment the model is prevented from making the costly mistake of predicting there are two tables instead of one.

In fact, no model predicts even a single table entirely correctly according to both content and structure, as ${Acc}_{Con}$ is 0 for all models.
The highest ${Acc}_{Top}$ is only 0.0625.
Thus, all models perform poorly on this task when taking the full set of metrics into account.

\subsection{Evaluating the Number of Tables Predicted Within a Page}
\label{sec:appendix_table_number_prediction}

\begin{table*}[tb]
\caption{The proportion of time a model predicts fewer tables (\colorbox{blue!25!white}{$<$}), equal number of tables (\colorbox{green!25!white}{$=$}), or more tables (\colorbox{red!25!white}{$>$}) compared to the ground truth number of tables for four types of pages in PubTables-v2 Full Documents: 1) pages with 0 tables, 2) pages with 1 table that is not split within the page, 3) pages with 1 table that is split within the page, and 4) pages with more than 1 table. Highest accuracy is bolded, highest error is underlined.}
\centering
\begin{adjustbox}{max width=\linewidth}
\begin{tabular}{l r r r r r r r r r r r}
\toprule
\multirow{2}{*}{\textbf{Model}} & \multicolumn{2}{c}{\textbf{0 tables}} & \multicolumn{3}{c}{\textbf{1 table, non-split}} & \multicolumn{3}{c}{\textbf{1 table, split}} & \multicolumn{3}{c}{\textbf{Multiple tables}} \\
\cmidrule(lr){2-3}\cmidrule(lr){4-6}\cmidrule(lr){7-9}\cmidrule(lr){10-12}
 & \multicolumn{1}{c}{\textbf{\colorbox{green!25!white}{$=$}}} & \multicolumn{1}{c}{\textbf{\colorbox{red!25!white}{$>$}}} & \multicolumn{1}{c}{\textbf{\colorbox{blue!25!white}{$<$}}} & \multicolumn{1}{c}{\textbf{\colorbox{green!25!white}{$=$}}} & \multicolumn{1}{c}{\textbf{\colorbox{red!25!white}{$>$}}} & \multicolumn{1}{c}{\textbf{\colorbox{blue!25!white}{$<$}}} & \multicolumn{1}{c}{\textbf{\colorbox{green!25!white}{$=$}}} & \multicolumn{1}{c}{\textbf{\colorbox{red!25!white}{$>$}}} & \multicolumn{1}{c}{\textbf{\colorbox{blue!25!white}{$<$}}} & \multicolumn{1}{c}{\textbf{\colorbox{green!25!white}{$=$}}} & \multicolumn{1}{c}{\textbf{\colorbox{red!25!white}{$>$}}} \\
\midrule
SmolDocling-256M & 0.989 & 0.011 & 0.193 & 0.769 & 0.038 & 0.103 & 0.676 & 0.221 & 0.186 & 0.722 & 0.092 \\
GraniteDocling-258M & 0.994 & 0.006 & 0.192 & 0.793 & 0.014 & 0.118 & 0.162 & 0.721 & 0.108 & 0.817 & 0.075 \\
Qwen2.5-VL-3b & \textbf{0.999} & 0.001 & \underline{0.275} & 0.361 & \underline{0.364} & \underline{0.250} & 0.176 & 0.574 & 0.176 & 0.359 & \underline{0.464} \\
Granite-Vision-3.2-2b & 0.357 & \underline{0.643} & 0.024 & 0.976 & 0.000 & 0.000 & \textbf{1.000} & 0.000 & \underline{1.000} & 0.000 & 0.000 \\
DeepSeek-OCR & 0.992 & 0.008 & 0.011 & 0.951 & 0.037 & 0.000 & 0.074 & 0.926 & 0.013 & 0.925 & 0.062 \\
DeepSeek-OCR 2 & 0.991 & 0.009 & 0.010 & 0.976 & 0.015 & 0.015 & 0.103 & 0.882 & 0.010 & 0.944 & 0.046 \\
dots.ocr & 0.990 & 0.010 & 0.008 & \textbf{0.978} & 0.014 & 0.015 & 0.044 & \underline{0.941} & 0.007 & \textbf{0.954} & 0.039 \\
\bottomrule
\end{tabular}
\end{adjustbox}
\label{tab:table_number_prediction}
\end{table*}

In this section, we analyze the results from the PubTables-v2 Full Documents experiment in \cref{sec:doc_level_te}, but looking only at how often the models produced the correct number of tables.
The results for this are given in \cref{tab:table_number_prediction}.
As can be seen, models differ in how often they over-predict or under-predict the true number of tables.
Failure to produce or predict the true number of tables helps to explain in part why certain models underperform relative to others according to the results in \cref{tab:experiment_multi_page_table_extraction}.

\subsection{Cross-Page Table Continuation Ablation Study}
\label{sec:cross_page_ablation}

\begin{table}[tb]
\caption{\textbf{Cross-page table continuation classification ablation study.} In this experiment, we show the impact that the number of unique training samples has on performance.}
\centering
\begin{tabular}{l r r r r}
\toprule
\textbf{Model} & \textbf{Samples} & \textbf{Recall} & \textbf{Precision} & \textbf{F1} \\
\midrule
ResNet-50 & 250 & 0.908 & 0.809 & 0.856 \\
 & 1,000 & 0.973 & 0.921 & 0.946 \\
 & 4,000 & 0.986 & 0.958 & 0.972\\
 & 15,830 & 0.986 & 0.973 & 0.979 \\
\midrule
ViT-B-16 & 250 & 0.902 & 0.824 & 0.862 \\
 & 1,000 & 0.983 & 0.914 & 0.947 \\
 & 4,000 & 0.988 & 0.972 & 0.980 \\
 & 15,830 & 0.995 & 0.987 & 0.991 \\
\bottomrule
\end{tabular}
\label{tab:experiment_cross_page_table_continuation_ablation}
\end{table}

The results of our cross-page table continuation experiment in \cref{sec:exp_cross_page_tables} show that standard image classification models trained on PubTables-v2 can learn with near perfect accuracy to predict if a table on one page continues onto the next.
In this experiment, we attempt to explore more deeply how difficult the task is by considering how the number of unique training examples impacts model performance.

The full training set contains 15,830 samples.
In addition to the full training set, we create 3 random subsets of size 250, 1000, and 4000.
We train each model for 200 epochs, evaluating on the validation set after each epoch and saving the model checkpoint that performs best according to F1 score.

The results of this experiment are presented in \cref{tab:experiment_cross_page_table_continuation_ablation}.
We can see clearly that model performance continues to improve with more data, suggesting that the most challenging cases do require a significant amount of data to be learned.

\subsection{Cross-Page Table Continuation Prediction Failures}
\label{sec:cross_page_failures}

\begin{figure}[tb]
\centering
    \includegraphics[width=0.85\linewidth]{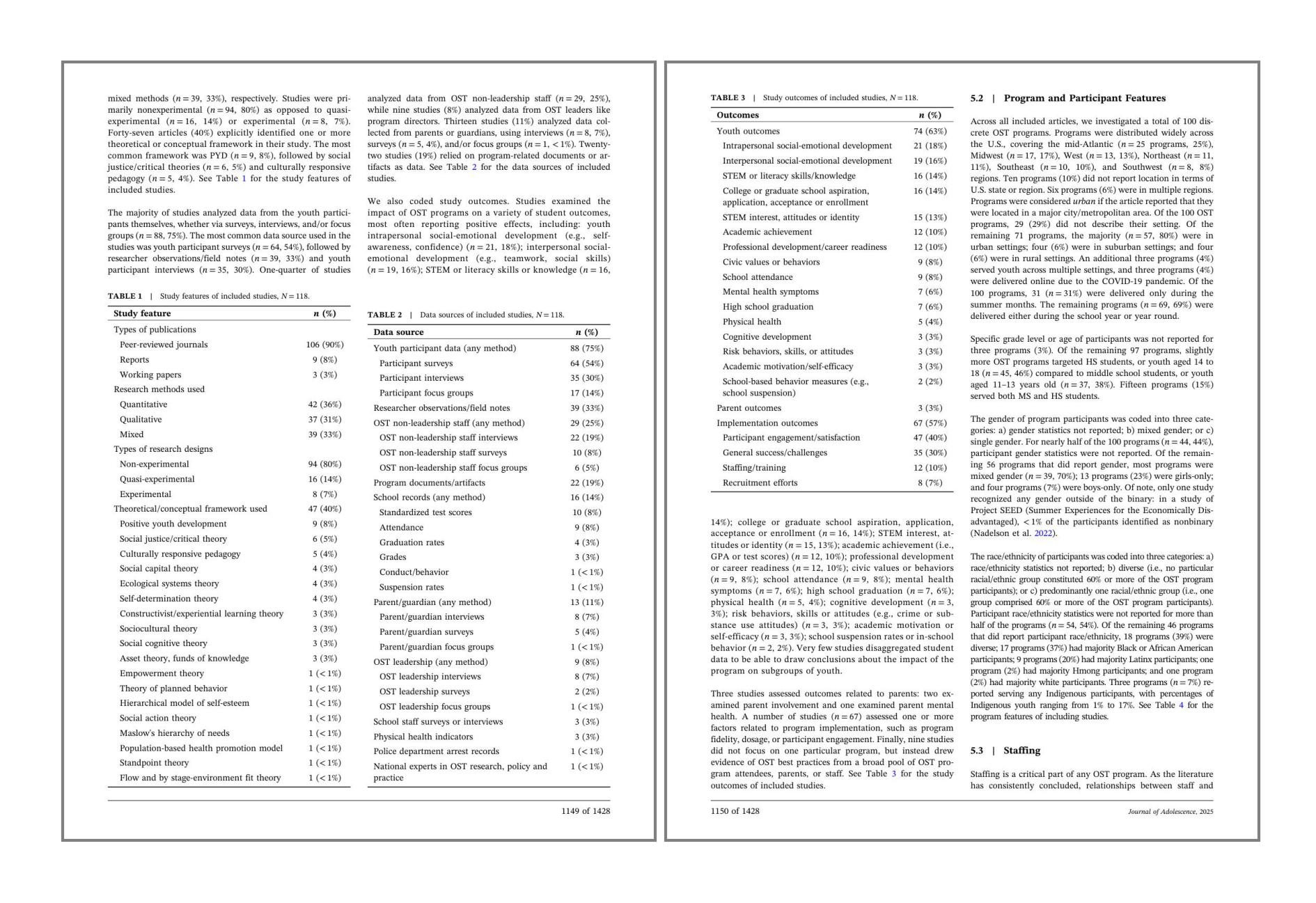}
    \caption{This figure shows one of the examples that both models from the cross-page table continuation prediction experiment fail on. Both models erroneously predict that the last table on the first page (left) continues onto the second page (right).}
    \label{fig:cross_page_failure_case}
\end{figure}

For the cross-page table continuation experiment in \cref{sec:exp_cross_page_tables}, the ViT model fails on 20 / 1862 cases, whereas the ResNet model fails on 46 / 1862 cases.
Both models fail on 16 / 1862 cases.
This suggests these 16 cases are the most difficult.
We show one of these failure cases in \cref{fig:cross_page_failure_case}.

In this example, both models erroneously predict that the table on the first page continues onto the second page.
In this case, multiple visual cues that the table could continue onto the next page are present, including the position of the tables and their similar content.

\subsection{VLM Qualitative Results}
\label{sec:appendix_vlm_results}

For Granite Vision, although we tried multiple prompting strategies, we never observed the model output more than one table, even when presented with a page containing multiple tables.
However, this does not mean that content from more than one table was never output.
For certain pages where two tables were close to each other and were similar sizes, both tables were mostly parsed and fused together into one output table.
Cases like this are penalized according to our metric and evaluation procedure, which matches at most one ground truth table to any predicted table.

For VLMs trained to parse documents into a specific format, we noticed that it is very important to follow the recommended format of the prompt to get results in the expected format.
In particular, for \texttt{Qwen2.5-VL-3B}, it is very important to include the recommended system prompt.
In our initial experiments, we discovered that if the system prompt is missing, the model does not produce QwenVL HTML consistently; instead, it produces the content of the page in a non-structured form.

\begin{figure}[tb]
\centering
    \includegraphics[width=0.75\linewidth]{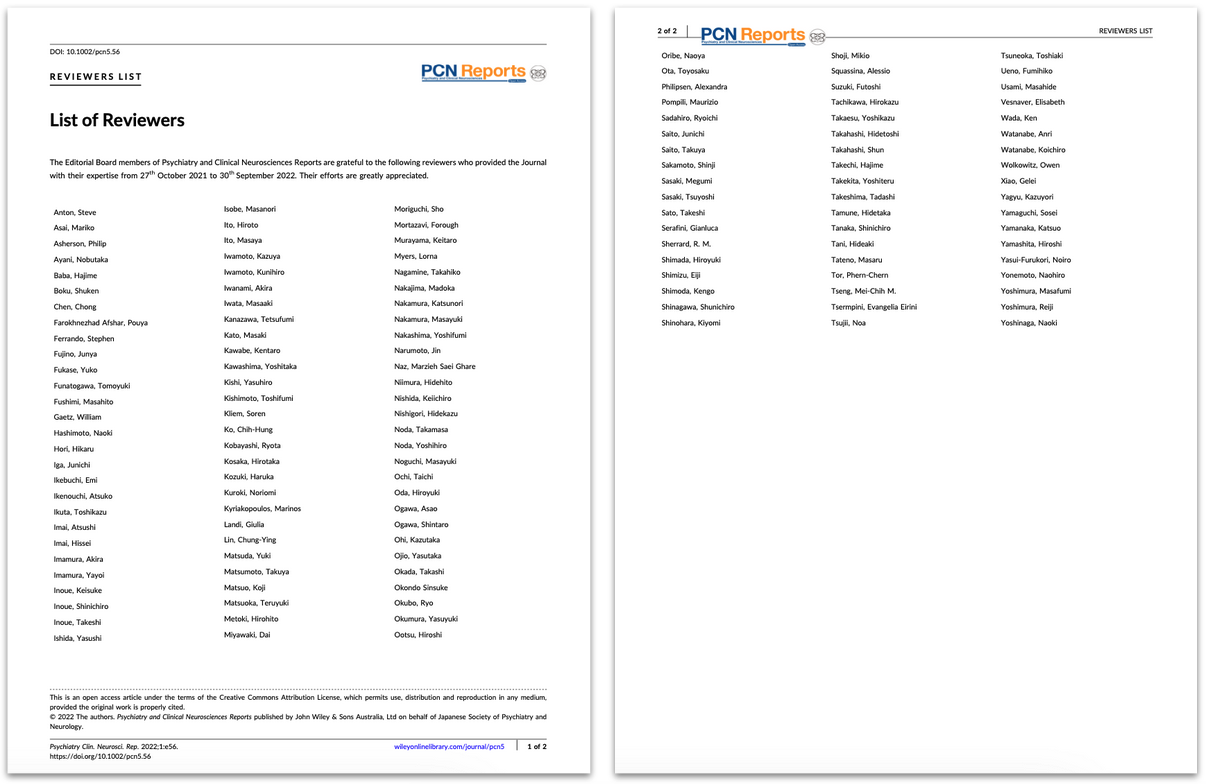}
    \caption{This figure shows the shortest document in the test set of the PubTables-v2 Full Documents collection, which has only two pages.}
    \label{fig:two_page_test_document}
\end{figure}

As noted earlier, table extraction from full documents remains a challenging task for the smaller, specialized VLMs we evaluate.
Among these models, only \texttt{Qwen2.5-VL-3B} is capable of processing almost all documents in the collection when given all pages at once, and its metrics are substantially lower compared to the results for the other two collections.

From a visual investigation of the generated text, we notice that a typical failure mode of \texttt{Qwen2.5-VL-3B} on this task is that it enters into an infinite loop repeating some portion of text from the document until it reaches the model's token limit.
Interestingly, this failure occurs even on the shortest document in the test set (shown in \cref{fig:two_page_test_document}), which is only two pages long.
We can see that this document lists names in a table-like form that spans two pages.
In this case, \texttt{Qwen2.5-VL-3B} ignores the table-like layout of the pages and lists all of the names as stand-alone HTML paragraphs.
It starts with the names on the first page and continues with the names on the second page.
However, when it reaches the final name on the second page, it enters into an infinite loop repeating the final name until it reaches the token limit.

We confirm that not all documents trigger this behavior.
We also confirm visually that this behavior can be observed with more conventional-looking PubMed articles from the test set.

Due to the computational resources required, we do not evaluate larger VLM models on the full test set.
However, we visually explore the predictions of a larger model, \texttt{Qwen2.5-VL-7B}, on several short documents.
\texttt{Qwen2.5-VL-7B} does not enter into an infinite loop on the document from \cref{fig:two_page_test_document}.
It captures the layout of the first page of the document, organizing the names in a three-column table.
However, it stops there, ignoring the second page completely.
We confirm that \texttt{Qwen2.5-VL-7B} can also enter into an infinite loop: both \texttt{Qwen2.5-VL-3B} and \texttt{Qwen2.5-VL-7B} generate text with repeating patterns until reaching the token limit for a four page document from the test set (albeit the exact patterns differ substantially between the models).
These observations present anecdotal evidence that the full document collection can be challenging even for larger VLMs.
A rigorous evaluation of their performance is beyond the scope of this paper.

\section{Ranking Stability Analysis}
\label{sec:appendix_ranking_stability}

\begin{figure}[tb]
    \centering
    \includegraphics[width=\linewidth]{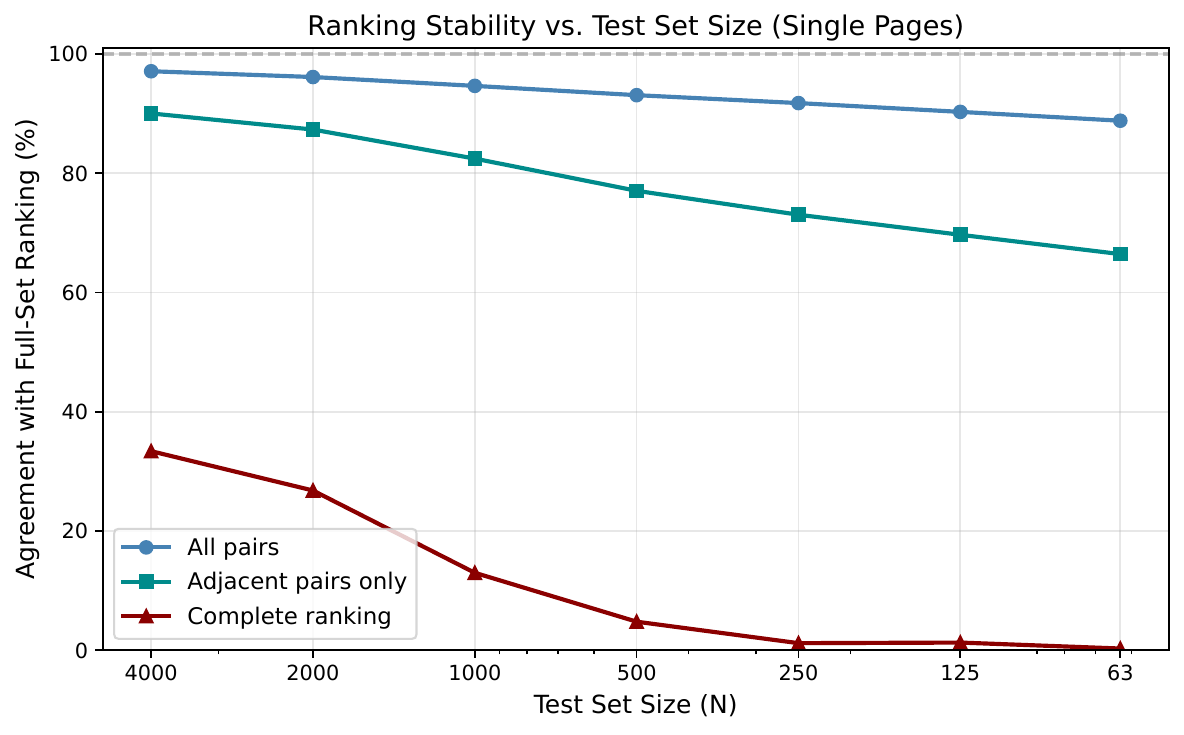}
    \caption{\textbf{Ranking stability vs.\ test set size} on Single Pages (1,000 bootstrap subsamples per $N$). Three progressively stricter metrics are shown: \emph{all pairs}---the fraction of (pair, subsample) combinations that agree with the full-set ordering; \emph{adjacent pairs}---restricted to consecutively-ranked models, where score gaps are smallest; and \emph{complete ranking}---the fraction of subsamples that preserve the entire model ordering simultaneously. While the all-pairs metric appears reassuring (${\sim}$90\% even at $N{=}63$), adjacent-pair agreement drops to 67\% and complete ranking collapses to $<$1\%, revealing that small test sets cannot reliably distinguish close models.}
    \label{fig:ranking_stability}
\end{figure}

To demonstrate that PubTables-v2's scale is necessary for reliable evaluation, we perform a bootstrap analysis measuring how model ranking stability depends on test set size.
We decompose GriTS$_\text{Con}$ scores on Single Pages (7,680 test samples) into per-sample sufficient statistics (true-positive score, number of ground-truth cells, number of predicted cells) and draw 1,000 subsamples of size $N$ without replacement for $N \in \{63, 125, 250, 500, 1000, 2000, 4000\}$.
This provides exact bootstrap distributions without re-running model inference.

We evaluate three progressively stricter stability metrics (\cref{fig:ranking_stability}).
\emph{All-pairs agreement}---the fraction of pairwise comparisons preserved---appears reassuring (90\% at $N{=}63$) but is dominated by easy comparisons between distant models.
\emph{Adjacent-pair agreement}, restricted to consecutively-ranked models, drops to 67\% at $N{=}63$ and only reaches 90\% at $N{=}4000$.
\emph{Complete ranking preservation} collapses to $<$1\% at $N{=}63$ and remains at only 33\% at $N{=}4000$.

This instability reflects genuine structural diversity within PubTables-v2, not merely small-sample noise.
If the test set were homogeneous, any random subset would yield nearly identical scores.
Instead, different models succeed or fail on different parts of the structural space---spanning cells, projected row headers, nested hierarchies, multi-table pages---and a small sample inevitably over-represents some patterns while missing others.
The full test set ensures this diversity is consistently represented, providing the statistical power needed for reliable model comparisons.
If a single domain already requires this scale, multi-domain benchmarks would likely need even larger test sets to achieve equivalent statistical reliability.

\end{document}